\documentclass[10pt,twocolumn,letterpaper]{article}

\usepackage[pagenumbers]{cvpr} 
\usepackage[normalem]{ulem}

\usepackage{amsmath}
\usepackage{amssymb}
\usepackage{mathtools}
\usepackage{amsthm}

\usepackage{enumitem}
\usepackage{wrapfig}
\usepackage{multirow}
\usepackage{listings}

\usepackage{booktabs}
\usepackage{multirow}
\usepackage{graphicx}

\usepackage{algorithm}
\usepackage{algpseudocode}

\usepackage{graphicx}
\useunder{\uline}{\ul}{}

\usepackage[table,xcdraw]{xcolor}

\theoremstyle{plain}

\newtheorem{lemma*}{Lemma}

\theoremstyle{definition}

\theoremstyle{remark}

\newcommand{\bu}{\mathbf{u}}
\newcommand{\bn}{\mathbf{n}}
\newcommand{\R}{\mathbb{R}}

\newcommand{\e}{\mathbf{e}}

\newcommand{\N}{\mathcal{N}}
\newcommand{\grad}{\nabla}
\newcommand{\x}{\mathbf{x}}
\newcommand{\g}{\mathbf{g}}
\newcommand{\rA}{\mathrm{A}}
\newcommand{\rB}{\mathrm{B}}
\newcommand{\rN}{\mathrm{N}}
\newcommand{\rG}{\mathrm{G}}

\definecolor{cvprblue}{rgb}{0.21,0.49,0.74}
\usepackage[pagebackref,breaklinks,colorlinks,citecolor=cvprblue]{hyperref}

\def\paperID{7569} 
\def\confName{CVPR}
\def\confYear{2025}

\title{Learning Normal Flow Directly From Event Neighborhoods}

\vspace{-21pt}
\author{
    Dehao Yuan, Levi Burner, Jiayi Wu, Minghui Liu, Jingxi Chen, \\Yiannis Aloimonos, Cornelia Fermüller\\
    University of Maryland, College Park, USA \\
    \href{https://github.com/dhyuan99/VecKM_flow}{https://github.com/dhyuan99/VecKM\_flow}
}

\begin{document}
\maketitle
\begin{abstract}
\vspace{-3pt}
Event-based motion field estimation is an important task. However, current optical flow methods face challenges: learning-based approaches, often frame-based and relying on CNNs, lack cross-domain transferability, while model-based methods, though more robust, are less accurate. To address the limitations of optical flow estimation, recent works have focused on normal flow, which can be  more reliably measured in regions with limited texture or strong edges. However, existing normal flow estimators are predominantly model-based and suffer from high errors.

In this paper, we propose a novel supervised point-based method for normal flow estimation that overcomes the limitations of existing event learning-based approaches. Using a local point cloud encoder, our method directly estimates per-event normal flow from raw events, offering multiple unique advantages: 1) It produces temporally and spatially sharp predictions. 2) It supports more diverse data augmentation, such as random rotation, to improve robustness across various domains. 3) It naturally supports uncertainty quantification via ensemble inference, which benefits downstream tasks. 4) It enables training and inference on undistorted data in normalized camera coordinates, improving transferability across cameras. Extensive experiments demonstrate our method achieves better and more consistent performance than state-of-the-art methods when transferred across different datasets. Leveraging this transferability, we train our model on the union of datasets and release it for public use. Finally, we introduce an egomotion solver based on a maximum-margin problem that uses normal flow and IMU to achieve strong performance in challenging scenarios.
\vspace{-2pt}
\end{abstract}

\vspace{-14pt}
\section{Introduction}
\vspace{-3pt}
\label{sec:introduction}
Event-based motion field estimation is a challenging task with significant potential for visual motion interpretation tasks, primarily due to its high temporal resolution, wide dynamic range, and low latency. Image motion is critical for various applications, such as egomotion estimation \cite{zhu2019unsupervised}, video interpolation \cite{chen2024timerewind, chen2024repurposing}, and motion deblurring \cite{xiong2024event3dgs}.

\begin{figure}
    \vspace{-0pt}
    \centering
    \vspace{-3pt}
    \includegraphics[width=0.95\linewidth]{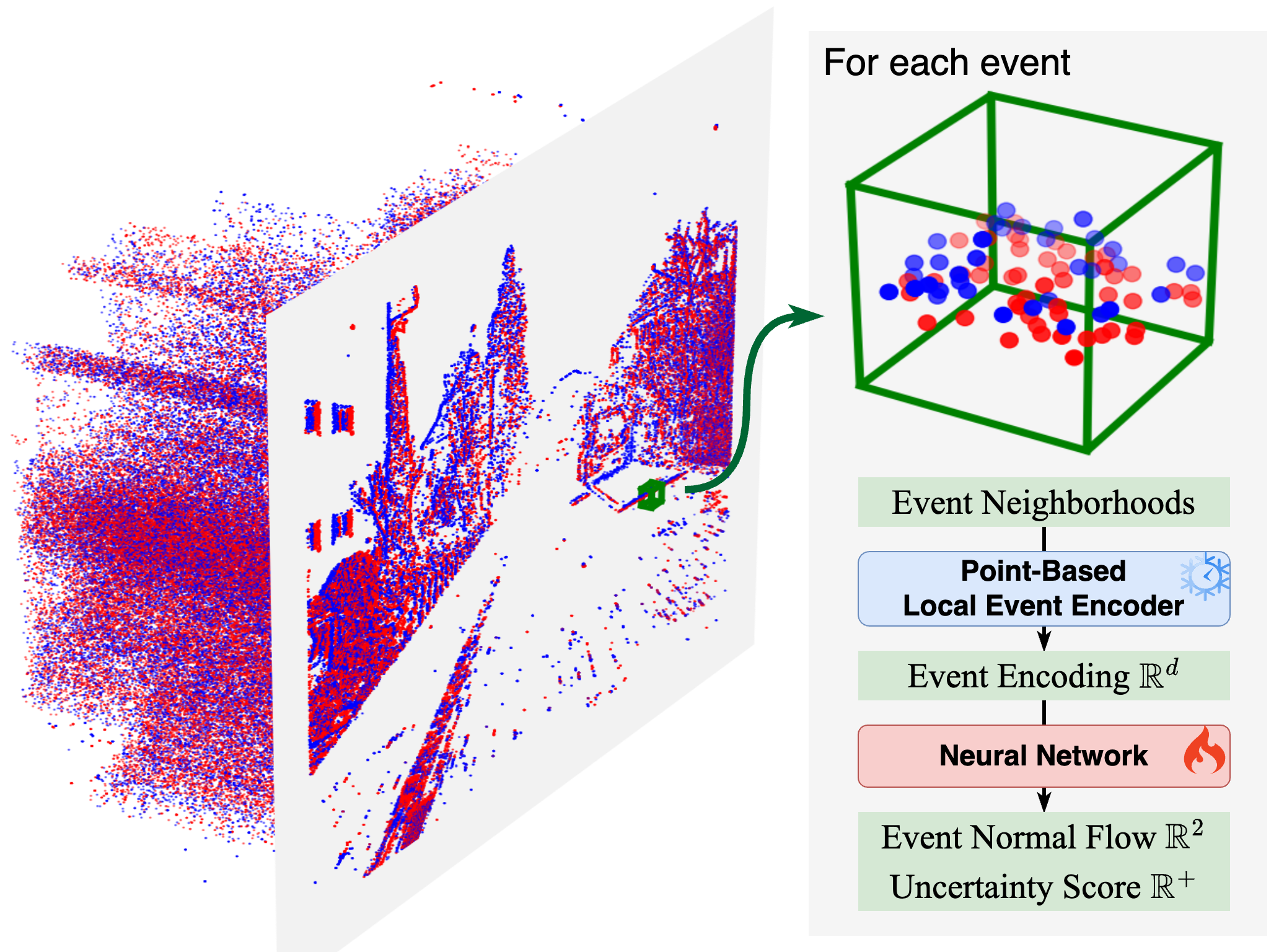}
    \caption{We propose a point-based network for estimating normal flow from raw event data. We discover multiple key advantages of this point-based approach compared with existing learning-based approaches. An event and its neighborhood are first encoded as a fixed-dimensional vector, which is then input to a network trained in a supervised way to predict normal flow. This approach achieves high accuracy while maintaining strong transferability across different domains and datasets. Besides, we demonstrate the usefulness of the estimated normal flow in a new egomotion solver that is shown to remain robust even during aggressive camera motions.}
    \label{fig:teaser}
    \vspace{-14pt}
\end{figure}

Most works on image motion focus on estimating optical flow (OF). Learning-based OF estimators like \cite{gehrig2021raft} perform impressively when evaluated within specific domains or datasets but suffer from accuracy degradation when applied across different domains, as shown in \cite{liu2023tma, li2023blinkflow}. In contrast, model-based OF estimators like \cite{shiba2022secrets} are more robust to domain shifts, but their accuracy is limited, particularly in scenarios where the event textures are sparse.

\begin{figure*}
    \centering
    \includegraphics[width=\linewidth]{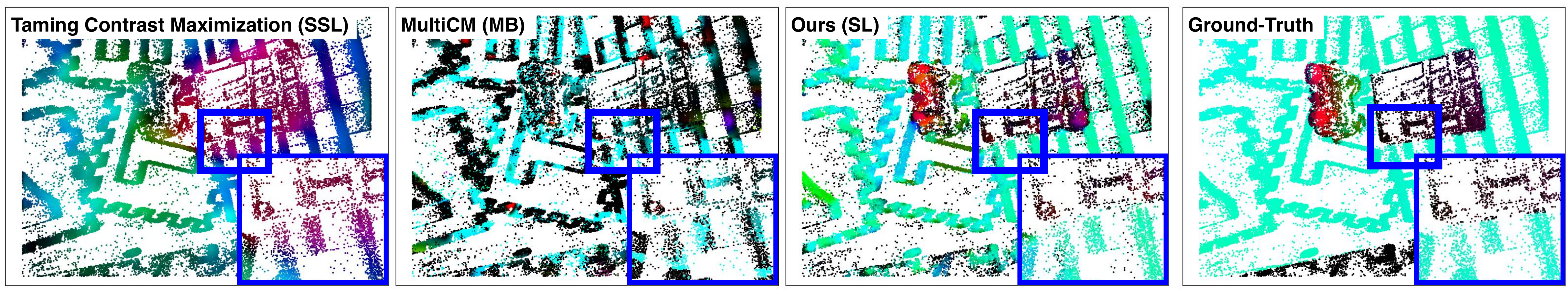}
    \vspace{-16pt}
    \caption{Our point-based method produces accurate and sharp predictions in the presence of independently moving objects, while other methods \cite{paredes2023taming, shiba2022secrets} fail. All models (if learning-based) are trained on DSEC and evaluated on EVIMO2. The flows are displayed in HSV color space, where the hue represents the flow direction, and the brightness represents the flow magnitude.}
    \label{fig:comparison}
    \vspace{-4pt}
\end{figure*}

The lack of robustness in optical flow estimators is mainly due to the local aperture problem \cite{local_aperture}. In regions with limited texture or strong linear edges, only the motion component perpendicular to the edges (normal flow) can be reliably measured, while the motion parallel to the edges remains ambiguous. To address this issue, optical flow estimators use CNN and RNN to enlarge their receptive fields, which makes the models vulnerable to overfitting. 

With this observation, some existing methods focus on predicting normal flow (NF) and demonstrate that normal flow is useful for tasks like egomotion estimation \cite{ren2024motion, lu2023event, li2024event}. However, current NF estimation approaches are predominantly model-based, relying on fitting a plane to the local space-time event surface \cite{benosman2013event,mueggler2015lifetime}. These approaches suffer from limited accuracy. Furthermore, the difficulty in obtaining ground-truth normal flow data  has impeded learning-based normal flow estimators.

As shown in Figure \ref{fig:teaser} and \ref{fig:comparison}, we explore supervised normal flow estimation while ensuring transferability across domains and datasets. Unlike most OF estimators that use CNNs on event frames, we propose a novel approach based on encoding 3D point sets and a novel loss function. This framework directly trains a network to estimate per-event normal flow from a local event cloud, supervised by ground-truth optical flow. To efficiently handle large event volumes, we adopt VecKM \cite{pmlr-v235-yuan24b}, a scalable and descriptive local point cloud encoder. We identify several unique advantages of estimating normal flow with this point-based network.
\begin{enumerate}
    \item \textbf{Temporally and Spatially Sharp Predictions}. 
    By predicting per-event flow based on the Euclidean neighborhood of every event, our method produces sharp predictions, especially for independently moving objects.
    \item \textbf{Richer data augmentation.} Our point-based approach enables a wider range of augmentations, such as uniform random rotations, which significantly enhances the estimator's accuracy and robustness across various domains.
    \item \textbf{Uncertainty quantification}. Our point-based method can compute prediction uncertainty with a simple ensemble inference, offering valuable insights for downstream tasks like egomotion estimation.
    \item \textbf{Strong transferability.} The method uses only event neighborhoods and we train on undistorted events in normalized camera coordinates. This improves transferability when training and testing on different datasets.
\end{enumerate}

\noindent Finally, we show the effectiveness of the predicted normal flows through aggressive egomotion estimation. We introduce a novel geometric-based egomotion solver that utilizes normal flows in conjunction with IMU measurements. This completes the pipeline integrating transferable event-based normal flow estimation with egomotion estimation. Our contribution can be summarized as follows:
\begin{itemize}
    \item We propose a learning-based, point-based normal flow estimator that offers multiple advantages over existing learning-based methods. It is more accurate than model-based normal flow estimators and more robust and transferable than learning-based optical flow estimators.
    \item We introduce a novel geometric-based egomotion solver only using normal flows and IMU measurements, which remains robust under aggressive egomotion scenarios.
    \item We extensively evaluate our point-based flow estimator on multiple datasets  and multiple transfer settings.
\end{itemize}




\section{Related Work}
\label{sec:related_work}
\subsection{Event-Based Optical Flow Estimation}
\label{sec:related_work_optical_flow}
In the early stages, \textbf{model-based} methods were studied for event-based optical flow estimation. The methodologies include extensions of the Lucas-Kanade algorithm \cite{benosman2012asynchronous,barranco2014contour}, feature matching \cite{liu2018adaptive, liu2022edflow, shiba2022fast, you2024vector}, contrast maximization (CM) \cite{gallego2018unifying}, Multi-CM \cite{shiba2022secrets}, plane fitting \cite{akolkar2020real, benosman2013event}, filter banks \cite{brosch2015event,barranco2015bio}, time surface matching \cite{almatrafi2020distance, brebion2021real, wan2024event}, iterative deblurring \cite{wu2024lightweight}.

Recently, \textbf{learning-based frame-based} methods, such as E-RAFT \cite{gehrig2021raft}, have dominated event-based optical flow estimation by leveraging correlation volumes. Many techniques, such as multi-modality \cite{zhu2018ev,wan2022learning,wan2023rpeflow, zhang2024cross}, motion aggregation \cite{liu2023tma, gehrig2024dense, paredes2023taming, ye2023towards}, synthetic datasets \cite{luo2023learning, li2023blinkflow, luo2024efficient} have improved these models. However, their performance drops significantly when tested across different domains.

Meanwhile, \textbf{spiking neural networks} (SNNs) are applied to event-based optical flow estimation due to their efficiency in processing asynchronous data \cite{NEURIPS2021_39d4b545, cuadrado2023optical, zhao2022learning, zhang2023event, ponghiran2023event, xu2024event}. SNNs offer energy efficiency and compatibility with neuromorphic hardware. However, their complex training and dependency on specialized hardware make them less practical than conventional neural networks.

Despite the existing research, \textbf{learning-based point-based} methods remain underdeveloped in the event camera community, mainly due to the challenges of training point-based estimators on large event datasets. Our work addresses this gap.

\subsection{Point-Based Networks for Event Cameras}
\label{sec:related_work_point}
Point-based networks have enabled direct processing of point cloud data where PointNet \cite{qi2017pointnet} was the pioneering work. Subsequently, many feature extractors were developed to improve point cloud processing \cite{qi2017pointnet++, zhao2021point, ma2022rethinking, zhang2023starting, wu2023mpct, yuan2023decodable}. All existing point-based networks follow a common pipeline that samples and groups input point clouds into centered neighborhoods, transforming the data from $(n, 3)$ to $(n, K, 3)$, where $K$ is the number of neighboring points. However, this method is impractical for event data due to the huge values of $n$ and $K$. Fortunately, with the invention of VecKM \cite{pmlr-v235-yuan24b}, a descriptive and scalable local point cloud encoder that eliminates the need for explicit grouping and sampling, it is now feasible to apply point-based networks to event data. We explore the potential in this paper.

Prior to the introduction of VecKM, several attempts were made to apply point-based networks to event-based vision tasks \cite{ren2024rethinking, sekikawa2019eventnet, wang2019space, yang2019modeling, ren2023spikepoint}. However, traditional point-based networks require downsampling of the local point cloud neighborhoods (i.e. reducing $K$), limiting their use to only high-level tasks like action recognition, which do not demand precise modeling of event geometry. No previous work has successfully used point-based networks for low-level tasks such as normal flow prediction, which requires an accurate representation of event geometry.
\subsection{Normal Flow and its Applications}
\label{sec:related_work_normal_flow}
Normal flow refers to the component of optical flow that is perpendicular to the edges or parallel to image gradients:
\begin{equation}
    \bn=-\frac{\nabla I\cdot \bu}{||\nabla I||^2}\grad I
\end{equation}
where $\grad I$ is the image gradient, $\bu$ is the optical flow vector, and $\bn$ is the normal flow. \textit{Normal flow can be estimated from a local neighborhood because it depends only on local spatial-temporal intensity changes.} Unlike optical flow, it is not affected by the aperture problem \cite{local_aperture}, as it only captures motion along the image gradient direction. Because normal flow is a projection of optical flow, it satisfies the following constraint, which will be used extensively in this paper:
\begin{equation}
    \bn\cdot(\bu-\bn)=0
    \label{eqn:normal_flow_constraint}
\end{equation}

Normal flow is typically estimated by fitting planes to very small event cloud neighborhoods \cite{benosman2013event,mueggler2015lifetime, pijnacker2018vertical, ren2024motion}, a simple and reliable method across various datasets. However, the approach encounters difficulties when the edge is curved, or the local region features a corner. 


Normal flow has been applied to motion and structure estimation \cite{ren2024motion, li2024event, lu2023event}. Traditionally,  3D motion has been computed from normal flow via classification approaches \cite{brodsky2000structure, fermuller1995passive} using the so-called depth positivity constraint, relating the 2D to the 3D measurements. Researchers recently developed ways to incorporate the depth positivity constraint into optimization frameworks \cite{barranco2021joint}, and neural networks \cite{parameshwara2022diffposenet} to estimate 3D motion.

The egomotion estimation algorithm proposed in this paper builds on the approach of \cite{barranco2021joint}, which also relies on the depth positivity constraint and optimization. However, we reformulate the optimization problem as training a support vector classifier, enhancing stability and accuracy.


\section{Methodology}
\subsection{Problem Definition and Overview}
\label{sec:problem_definition}
\textbf{Point-Based Normal Flow Prediction}. The input to our point-based normal flow estimator is a sequence of events $(\e_1, \e_2, \cdots, \e_N)$, where $\e_k=(t_k, x_k, y_k)$\footnote{We omit polarity here because we found the polarity does not improve prediction accuracies. 
}. 
We assume access to GT per-event optical flow $(\bu_1, \bu_2, \cdots, \bu_N)$ for supervision (see Appendix \ref{app:dataset_preprocess} for how to obtain it). 
The output of the estimator is a sequence of normal flow predictions $(\hat{\bn}_1, \hat{\bn}_2, \cdots, \hat{\bn}_N)$, where $\hat{\bn}_k\in\R^2$. Each prediction $\hat{\bn}_k$ is determined by the centered neighboring events of $\e_k$:
\begin{align}
\begin{split}
    &\hat{\bn}_k = f\big(\N(\e_k)\big) \text{, where } \N(\e_k):=\\
  &\Big\{\e_j-\e_k: \Big|\Big|\Big(\frac{t_j-t_k}{\delta t}, \frac{x_j-x_k}{\delta x}, \frac{y_j-y_k}{\delta y}\Big)\Big|\Big|_{2} < 1\Big\}  
\end{split}
\label{eqn:local_events}
\end{align}
where $\delta t$, $\delta x$, $\delta y$ are the hyper-parameters controlling the neighborhood size. Note we allow the number of neighboring events to be different, i.e., $|\N(\e_k)|$ can be different for different events $\e_k$. As discussed in Section \ref{sec:related_work_normal_flow}, the neighboring events contain sufficient information to estimate normal flow because it is determined by intensity gradients, which are inherently local.

\begin{figure}
    \centering
    \includegraphics[width=0.91\linewidth]{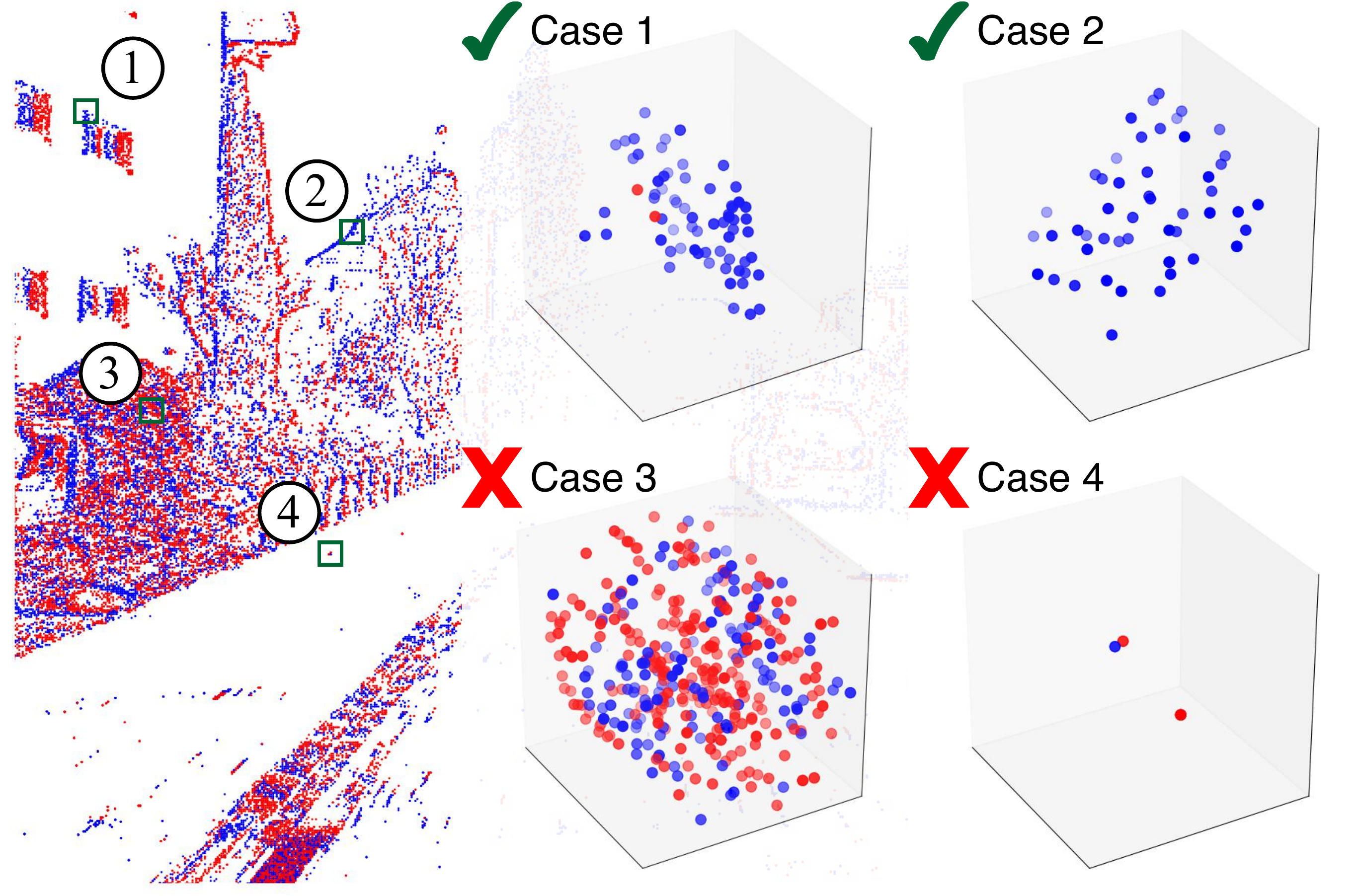}
    \caption{
    Uncertainty quantification (UQ) is important for per-event normal flow estimation, as it helps filter out less reliable predictions. For example,  the normal flow predictions in Cases 1 and 2 are more reliable compared to those in Cases 3 and 4.}
    \label{fig:important_uq}
    \vspace{-10pt}
\end{figure}

\noindent\textbf{Uncertainty Quantification (UQ)}. After predicting the per-event normal flows $(\hat{\bn}_k)_{k=1}^N$, we estimate uncertainty scores $(\sigma_k)_{k=1}^N$ of the predictions, where $\sigma_k \geq 0$.\hspace{-0.5pt} UQ is crucial in per-event normal flow estimation because, as shown in Figure \ref{fig:important_uq}, there are  neighborhoods where normal flow can not be estimated reliably. UQ identifies and removes these noisy predictions before passing them to downstream tasks.

\noindent\textbf{Overview}. Our solution to the problem is outlined in Figure \ref{fig:teaser}. Each component of the pipeline is detailed as followed. Section \ref{sec:local_events} introduces the local events encoder, which efficiently transforms the local events $\N(\e_k)$ into a representative vector. Section \ref{sec:motion_field_loss} introduces a novel loss function that guides the network to predict normal flows, supervised by optical flow ground-truth. Section \ref{sec:data_augmentation} introduces the data augmentation tricks used during training, which is a unique advantage of our point-based method. Section \ref{sec:uncertainty_quantification} introduces how to compute uncertainty scores during inference time. Section \ref{sec:egomotion_estimation} details an egomotion estimation algorithm that uses normal flow inputs, where we enhance the robustness and stability of the algorithm from \cite{barranco2021joint}.

\subsection{Local Events Encoder}
\label{sec:local_events}
We introduce how to encode the neighboring events $\N(\e_k)$ into a vector representation for each event $\e_k$. Given the high volume of input events($\sim80$k events every 20 ms), the local events encoder must be scalable and efficient. 

We use a recently developed a local geometry encoder named VecKM \cite{pmlr-v235-yuan24b}, which is designed to process large point clouds efficiently. VecKM models local events as samples from a kernel mixture and uses random Fourier features to transform the kernel mixture into a vector representation. Such formulation allows using all neighboring points to compute the local geometric encoding without down-sampling the neighborhood. Besides, it eliminates the need for explicitly grouping and sampling the event neighborhoods. These merits make VecKM well-suited for handling high-volume event data and dense local event regions.

\noindent\textbf{VecKM Encoding of Local Events.} Given the normalized events $X_{N\times 3}=\{(\frac{t_k}{\delta t}, \frac{x_k}{\delta x}, \frac{y_k}{\delta y})\}_{k=1}^N$ in Eqn. \eqref{eqn:local_events}, the VecKM local events encoding $G_{N\times d}$ is computed by:
\begin{align}
\begin{split}
    J_{N\times N}&=adjacency\_matrix(X_{N\times 3}) \\
    \mathcal{A}_{N\times d}&=\exp(iX_{N\times 3} A_{3\times d}) \\
    G_{N\times d}&=normalize\big((J_{N\times N} \mathcal{A}_{N\times d}) ./ \mathcal{A}_{N\times d}\big)
\end{split}
\label{eqn:VecKM_encoding}
\end{align}
$J$ is a sparse adjacency matrix where the $(j,k)$-entry is 1 if $\e_j$ and $\e_k$ are close, and 0 otherwise. $A_{3\times d}$ is a randomized fixed matrix with entries drawn from a normal distribution with a mean of 0 and variance of 25. The function $\exp(i\cdot)$ is the element-wise Euler formula, and $./$ denotes the element-wise complex number division.

The neighbor information is captured in the sparse adjacency matrix $J$ and incorporated into the local event encoding $G$ through matrix multiplication. The centralizing step is implicitly achieved by the element-wise division. This formulation eliminates the need for explicitly grouping and sampling event neighborhoods, making it descriptive and scalable. Kindly refer to \cite{pmlr-v235-yuan24b} for detailed derivation.

\begin{figure}
    \centering
    \includegraphics[width=0.9\linewidth]{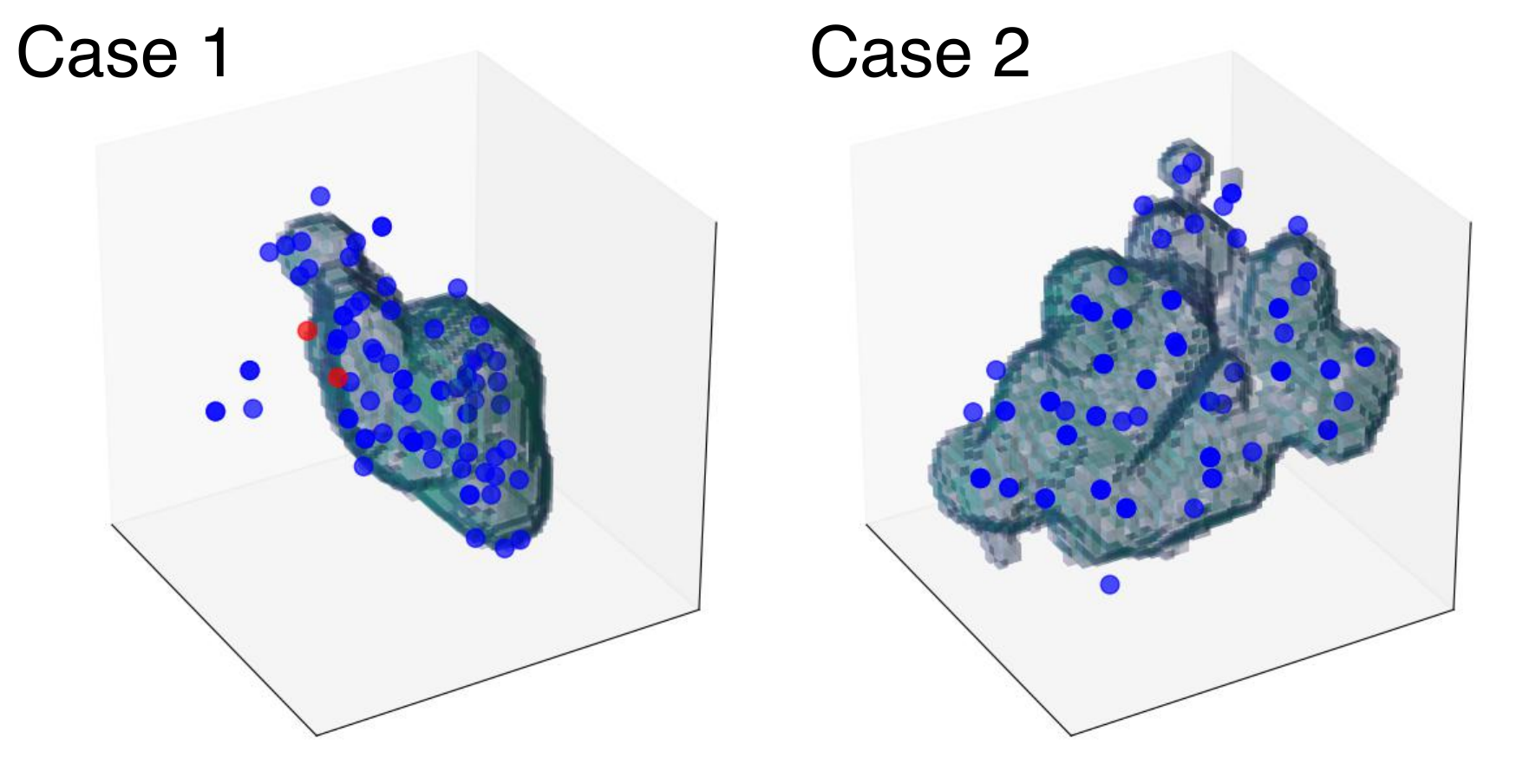}
    \caption{Reconstruction of a density distribution (shown in gray) from VecKM's local events encoding. The reconstructed 3D distribution closely aligns with the original (blue and red) events, demonstrating that VecKM's encoding effectively represents the event data. The examples shown are identical to those in Figure \ref{fig:important_uq}.}
    \label{fig:qual_VecKM}
\end{figure}

\noindent\textbf{Qualitative Evaluation of VecKM Encoding.} \hspace{-1pt}VecKM \cite{pmlr-v235-yuan24b} proves that $G[k,:]\in\mathbb{C}^d$ in Eqn. \eqref{eqn:VecKM_encoding} effectively represents local events $\N(\e_k)$. Specifically, the local events distribution can be reconstructed from the encoding $G[k,:]$. Figure \ref{fig:qual_VecKM} shows examples of local events and the distribution reconstruction from the local events encoding. The strong alignment between the events and the reconstructed distributions suggests that VecKM generates a representative encoding.

\noindent\textbf{Normal Flow Prediction.} After computing the local events encoding $G_{N\times d}$ from Eqn. \eqref{eqn:VecKM_encoding}, we transform each event's encoding $G[k,:]$ to normal flow prediction $\hat{\bn}_k$ using a multi-layer perceptron (MLP) \cite{trabelsi2017deep}. The MLP is trained by a novel loss function introduced in the next section.

\subsection{Normal Flow Learning with a Two-Term  Loss}
\label{sec:motion_field_loss}

\begin{figure}
    \centering
    \includegraphics[width=\linewidth]{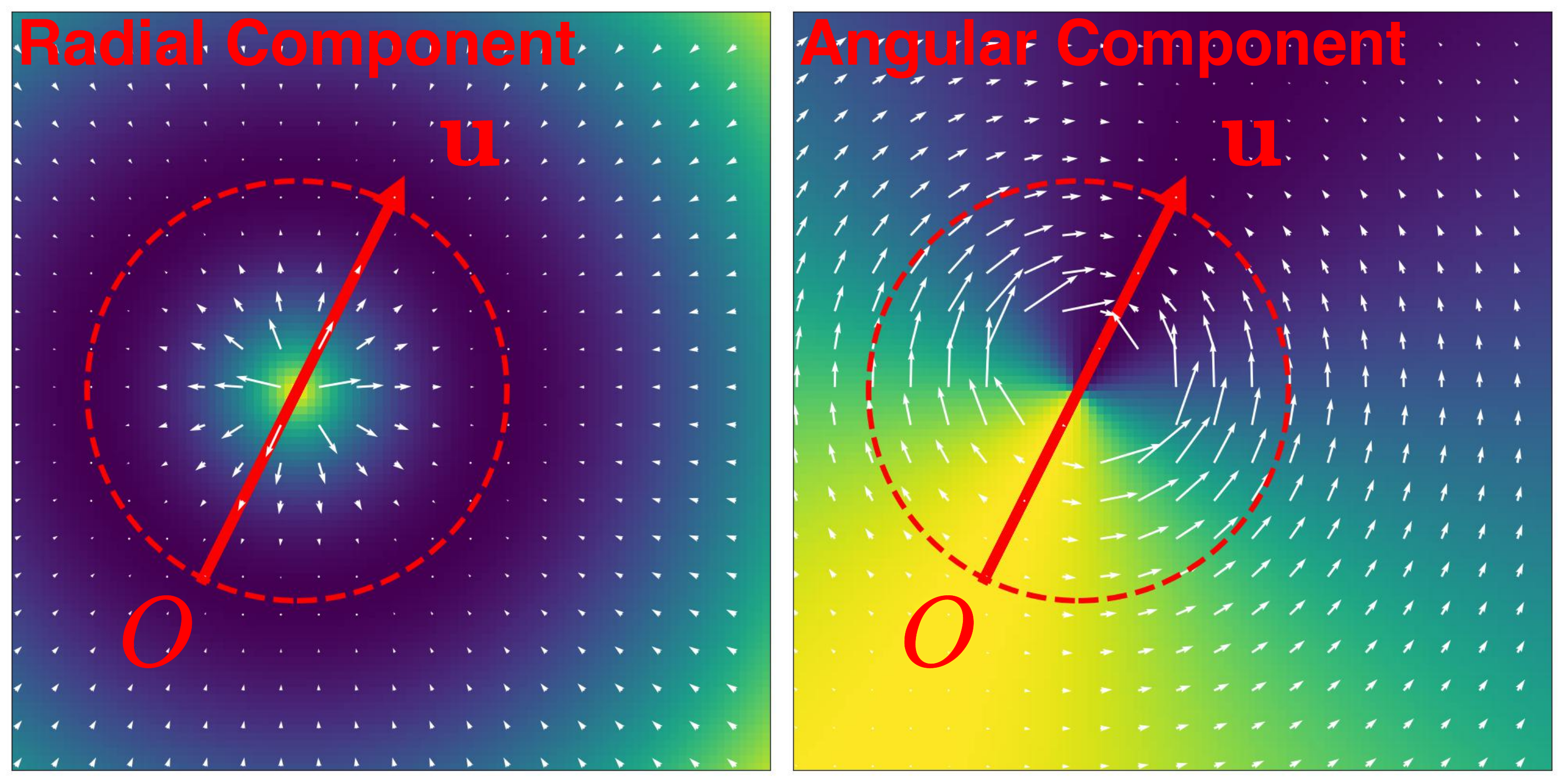}
    \caption{Loss maps and gradient fields of the motion field loss function. Our motion field loss function consists of radial and angular components. Given the GT optical flow $\bu$, the radial component guides the predicted flow to lie on the circle with $\bu$ as the diameter. The angular component guides the predicted flow to align with $\bu$, which prevents the trivial prediction of zero flow.}
    \label{fig:motion_field_loss}
    \vspace{-12pt}
\end{figure}

In frame-based vision, {\it normal flow} has been defined as the flow component along the image gradient (or perpendicular to the local edge). In event space, however, no obvious definition exists for an image gradient that can be computed per-event
and so obtaining ground truth (GT) normal flow for an event camera is challenging. Instead, we use GT optical flow as supervision, and guide the network to predict normal flow (the component of optical flow along some direction) by training with a novel motion field loss function. This loss function is composed of two components: a radial and an angular component, as illustrated in Figure \ref{fig:motion_field_loss}.


The \textbf{radial component} guides the predicted flow to satisfy the normal flow magnitude constraint \eqref{eqn:normal_flow_constraint}. Geometrically, this means the predicted flow lies on a circle where the GT optical flow represents the diameter. When the normal flow constraint is satisfied, the radial loss becomes zero.

Note that predicting zero flow is a trivial solution for minimizing the radial component. To avoid such prediction, we introduce an \textbf{angular component}, which encourages the predicted flow to have the same direction as the optical flow.

Heuristically, training on the sum of these two components allows the network to predict the full optical flow when possible. If the texture information is insufficient to predict the full optical flow, the network resorts to predicting the normal flow. We show that our intention is achieved in Figure \ref{fig:effect_motion_field} in Appendix \ref{app:motion_field_loss}.

Mathematically, given the GT optical flow, the normal flow is predicted using the following loss function:
\begin{align}
    Radial(\bu, \hat{\bn}) &= \log\Big(\frac{\epsilon+||\hat{\bn}-\bu/2||}{\epsilon+||\bu/2||}\Big)^2 \label{eqn:radial}\\
    Angular(\bu, \hat{\bn}) &= -\frac{(\hat{\bn}-\bu/2)\cdot\bu}{||\hat{\bn}-\bu/2||\cdot ||\bu||}
    \label{eqn:motion_field_loss}
\end{align}
Figure \ref{fig:motion_field_loss} shows the resulting loss map and gradient field. Note, that the two components gradients are orthogonal everywhere, so the optimization of one component does not interfere with the other.

\subsection{Improve Transferability by Data Augmentation}
\label{sec:data_augmentation}
Diverse data augmentation is a unique advantage of our point-based network, which improves the transferability of the estimator. Below, we enumerate the data augmentation tricks applied during training. We denote the input events as $X_{N\times 3}$ and their corresponding GT optical flow as $U_{N\times 2}$.

\noindent\textbf{Random Rotation}. Motion field estimation is an equivariant task, meaning that rotating the events on the image plane causes the motion field to rotate by the same angle. Thanks to this, we apply a uniformly sampled random rotation angle $\theta\in[0, 2\pi)$ to the event cloud and GT flow. The augmented inputs and targets are obtained by:
\begin{align}
\begin{split}
    X_{N\times 3}&\leftarrow X_{N\times 3}\left[\begin{matrix} 1 & 0 & 0 \\ 0 & \cos\theta & -\sin\theta \\ 0 & \sin\theta & \cos\theta
    \end{matrix}\right]  \\
    U_{N\times 2}&\leftarrow U_{N\times 2}\left[\begin{matrix}\cos\theta & -\sin\theta \\ \sin\theta & \cos\theta \end{matrix}\right]
\end{split}
\label{eqn:random_rotation}
\end{align}
This forces the network to estimate direction without bias.

\noindent\textbf{Random Scaling}. We scale the event streams by a random scalar $\alpha \in (0.75, 1.25)$. The augmented inputs and targets are obtained by: $X \leftarrow X * \alpha$, $U \leftarrow U$. This improves the estimator's performance on small and large objects.

\noindent\textbf{Random Sampling}. We sample 50\% $\sim$ 100\% percentage of the events and their corresponding flows.
This makes the estimator robust to density variation.

\subsection{Inference with Uncertainty Quantification}
\label{sec:uncertainty_quantification}
As explained in Section \ref{sec:data_augmentation} \textbf{Random Rotation}, motion field estimation is an equivariant task, which we exploit to estimate an uncertainty score based on how well the predicted flow remains equivariant to rotation. Specifically, we sample $K$ rotation angles and infer the normal flow with the rotated events. Then we de-rotate the predicted normal flow with the corresponding rotation angle:
\begin{align}
    \hat{U}&= estimator\Big(X_{N\times 3}\left[\begin{matrix}
        1 & \mathbf{0}^T \\ \mathbf{0} & R(\theta)
    \end{matrix}\right]\Big) R(\theta)^{-1}
    \label{eqn:ensemble}
\end{align}

By doing so, we obtain an ensemble of $K$ predicted flows for each event, and can test their consensus to compute an uncertainty score. Specifically, we use the circular standard deviation \cite{circstd} of the ensemble as the uncertainty score. The final prediction is given by the average of the ensemble in polar coordinates. Predictions corresponding with uncertainty above a threshold are discarded.

\subsection{Egomotion Estimation from Normal Flow}
\label{sec:egomotion_estimation}
We demonstrate the usefulness of our normal flow for egomotion estimation. We assume within a short time interval, we have normalized the event coordinates, per-event normal flow predictions, and we have a rotational velocity estimate from the IMU sensor. We solve for the translation direction.

\hspace{-2pt}Our egomotion solver uses the depth positivity constraint, which states that all world points are in front of the camera, i.e., have positive depth. Existing methods \cite{fermuller1995passive, barranco2021joint, parameshwara2022diffposenet} utilize this constraint by maximizing the negative depths.
However, the solution to this problem is not unique with respect to normal flow because depth positivity only considers the sign (or the direction) of the normal flow and is thus a weak constraint.
So, when estimated normal flow is used, errors in the estimated normal flow direction determine the output. Thus we reformulate the problem as maximum margin problem solved by training a support vector classifier, which results in a robust estimate.

Given an estimate of the angular velocity $\Omega$ from the IMU, we can calculate the rotational component $\bu^{rot}_\x = B_\x \Omega$ of the optical flow at pixel $\x=(x,y)$ where $B_\x$ is:
\begin{align}
\begin{split}
    B_\x=&\left[\begin{matrix}
        xy & -(x^2+1) & y \\
        (y^2+1) & -xy & -x
    \end{matrix}\right] 
\end{split}
\label{eqn:B}
\end{align}
Then the magnitude of the derotated normal flow at pixel $\x$, called $n_\x$, can be calculated through the relation 
$n_\x = \|\hat{\bn}_\x\| - \g_\x^T \bu^{rot}_\x$ where $\g_\x\in\R^{2\times 1}$ is the direction of normal flow with unit norm.
Then $n_\x$ is related to translation by:
\begin{equation}
    n_\x = \frac{1}{Z_\x}(\g_\x^T A_\x)V \label{eqn:egomotion_constraint}
\end{equation}
Where $V\in\R^{3\times1}$ denotes the direction of translation, $Z_\x\in\R$ denotes depth, and $A_\x$ is:
\begin{align}
\begin{split}
    A_\x=&\left[\begin{matrix}
        -1 & 0 & x \\ 0 & -1 & y
    \end{matrix}\right] \\ 
\end{split}
\label{eqn:AB}
\end{align}
Since the depth $Z_\x$ is positive, the following product, denoted as $\rho_\x(V)$, should be positive for all $\x$:
\begin{equation}
    \rho_\x(V)=n_\x(\g_\x^TA_\x V)   > 0
    \label{eqn:rho}
\end{equation}
We impose this constraint by formulating Eqn. \eqref{eqn:rho} into a linear support vector classification without intercept, namely solving $V$ such that $(\g_\x^T A_\x) V$ has a target sign decided by $n_\x$.
The pseudo-code is given below. $p$ is the number of events. $\rA$, $\rB$ are obtained through Eqn. \eqref{eqn:B} and \eqref{eqn:AB}, $\rN$ are the norms of the predicted normal flows, $\rG$ are the directions of the predicted normal flows with unit lengths, and $\Omega_0$ is the rotational estimate from the IMU sensor.

\begin{algorithm}
\caption{Egomotion Solver}\label{alg:example}
\begin{algorithmic}[1]
\State \textbf{Input:} $\rA_{p\times 2\times 3}$, $\rB_{p\times 2\times 3}$, $\rN_{p\times 1}$, $\rG_{p\times 2}$, $\Omega_0\in\mathbb{R}^{3\times 1}$.
\State \textbf{Output:} $V\in\mathbb{R}^{3\times 1}$.
\State $Q_{p\times 3} =$ \texttt{batch\_mat\_mul}$(\rG, \rA)$
\State $R_{p\times 1} =$ $\rN - $ \texttt{batch\_mat\_mul}$(\rG, \rB)\times \Omega_0$
\State $Q_{2p\times 3}=$ \texttt{concat($Q$,$-Q$)}
\State $R_{2p\times 1}=$ \texttt{sign(concat($R$,$-R$))}
\State \texttt{svm = LinearSVM(fit\_intercept=False)}
\State $V_{3\times 1} = $\texttt{ svm.fit($Q$,$R$).coef\_}
\State $V = V \:/\: ||V||$
\end{algorithmic}
\end{algorithm}

\vspace{-7pt}
\section{Experiments on Normal Flow Estimation}
\label{sec:s40_experiments}
\noindent\textbf{Datasets.} We use MVSEC \cite{zhu2018multivehicle}, EVIMO2 \cite{burner2022evimo2}, and  DSEC \cite{Gehrig3dv2021} to evaluate the accuracy and transferability of our normal estimator. We undistort and transform the events to normalized camera coordinates (focal length one). We train on each dataset and evaluate the model on all three datasets, resulting in nine combinations. Additionally, we train our model on the union of the three (and potentially more) datasets. We evaluate it and release it for public use.

Note that MVSEC has different camera resolution and distortion compared to EVIMO2 and DSEC. Therefore, there is a substantial domain gap among the cameras used in the three datasets.




\noindent\textbf{Evaluation Metrics.} Quantitatively, we use projection endpoint error (PEE) and percentage of sign correctness ($\% Pos$), following the convention in \cite{parameshwara2022diffposenet}, to evaluate the accuracy of normal flow predictions. 
\begin{align}
    PEE(\bu, \hat{\bn})&=\bigg|\bigg|\frac{\bu\cdot \hat{\bn}}{||\hat{\bn}||} - ||\hat{\bn}||\bigg|\bigg| \\ 
    \% Pos(\bu, \hat{\bn}) &= percentage(\bu\cdot\hat{\bn}>0)
\end{align}
PEE measures how well the normal flow magnitude constraint \eqref{eqn:normal_flow_constraint} is satisfied. It measures the error in the length of the normal flow.  $\%Pos$ measures how well the predicted normal flow has correct orientation as determined by the GT optical flow. Note that since the direction of normal flow is independent of motion, optical flow also satisfies the constraint equation \eqref{eqn:normal_flow_constraint} and thus also minimizes PEE.

\noindent\textbf{Qualitative Evaluation}. In \href{https://drive.google.com/drive/u/3/folders/1gkmUyZX5VRf8DxiBKL9CSdWdifjqZVq3}{this drive}, we include flow prediction videos for every evaluation. Qualitative evaluation is crucial for assessing aspects like prediction sharpness and handling independently moving objects, which are not fully captured by quantitative metrics. We strongly encourage readers to focus on the qualitative results for a comprehensive assessment of model performance. We show visualizations in Figure \ref{fig:EVIMO_vis} and upload flow prediction videos to \href{https://drive.google.com/drive/u/3/folders/1gkmUyZX5VRf8DxiBKL9CSdWdifjqZVq3}{this drive}. See Appendix \ref{app:flow_prediction_video} for detailed video descriptions.

\noindent\textbf{Compared Models.} We compare our point-based learning-based normal flow estimator with state-of-the-art event-based optical flow estimators.
We compare against MultiCM \cite{shiba2022secrets}, E-RAFT \cite{gehrig2021raft}, and Taming Contrast Maximization (TCM) \cite{paredes2023taming}, which are all frame-based estimators using model-based, supervised, and self-supervised learning approaches. Additionally, we compare against two point-based normal flow estimators: PCA 
\cite{benosman2013event} and PointNet \cite{qi2017pointnet}.

\noindent\textbf{Implementation details and hyper-parameter setting} are presented in Appendix \ref{app:our_model}.

\subsection{Evaluation on MVSEC}
\label{sec:s41_MVSEC}
\begin{table*}[t]
\centering
\resizebox{0.8\textwidth}{!}{%
\begin{tabular}{lccccccccccccc}
\hline
\multicolumn{2}{l}{} &
   &
   &
  \multicolumn{2}{c}{Indoor Flying 1} &
  \multicolumn{2}{c}{Indoor Flying 2} &
  \multicolumn{2}{c}{Indoor Flying 3} &
  \multicolumn{2}{c}{Outdoor Day 1} &
  \multicolumn{2}{c}{Average} \\ \cline{5-14} 
\multicolumn{2}{l}{\multirow{-2}{*}{}} &
  \multirow{-2}{*}{Input} &
  \multirow{-2}{*}{\begin{tabular}[c]{@{}c@{}}Training\\ Set\end{tabular}} &
  PEE $\downarrow$ &
  \% Pos $\uparrow$ &
  PEE $\downarrow$ &
  \% Pos $\uparrow$ &
  PEE $\downarrow$ &
  \% Pos $\uparrow$ &
  PEE $\downarrow$ &
  \% Pos $\uparrow$ &
  PEE $\downarrow$ &
  \% Pos $\uparrow$ \\ \hline
MultiCM &
  MB &
  F &
  - &
  \cellcolor[HTML]{F4C5C0}0.993 &
  \cellcolor[HTML]{AEDEC7}{\ul 97.8\%} &
  \cellcolor[HTML]{F7D1CD}1.378 &
  \cellcolor[HTML]{A9DCC4}{\ul 98.1\%} &
  \cellcolor[HTML]{F3BEBA}1.191 &
  \cellcolor[HTML]{A5DBC1}{\ul 98.2\%} &
  \cellcolor[HTML]{EA8F88}1.422 &
  \cellcolor[HTML]{FFFFFF}93.2\% &
  \cellcolor[HTML]{F2B7B2}1.246 &
  \cellcolor[HTML]{B8E3CE}{\ul 96.8\%} \\
PCA &
  MB &
  P &
  - &
  \cellcolor[HTML]{ED9D96}1.460 &
  \cellcolor[HTML]{FFFFFF}76.2\% &
  \cellcolor[HTML]{F4C5C1}1.586 &
  \cellcolor[HTML]{FFFFFF}76.3\% &
  \cellcolor[HTML]{ED9F98}1.552 &
  \cellcolor[HTML]{FFFFFF}75.0\% &
  \cellcolor[HTML]{E67C73}1.548 &
  \cellcolor[HTML]{FFFFFF}79.9\% &
  \cellcolor[HTML]{EC9A93}1.537 &
  \cellcolor[HTML]{FFFFFF}76.9\% \\ \hline
 &
   &
   &
  M &
  \cellcolor[HTML]{E67C73}1.836 &
  \cellcolor[HTML]{FFFFFF}76.1\% &
  \cellcolor[HTML]{E67C73}2.867 &
  \cellcolor[HTML]{FFFFFF}74.7\% &
  \cellcolor[HTML]{E67C73}1.951 &
  \cellcolor[HTML]{FFFFFF}75.4\% &
  \cellcolor[HTML]{FFFFFF}\textbf{0.677} &
  \cellcolor[HTML]{B1E0C9}{\ul 97.8\%} &
  \cellcolor[HTML]{E67C73}1.833 &
  \cellcolor[HTML]{FFFFFF}81.0\% \\
\multirow{-2}{*}{E-RAFT} &
  \multirow{-2}{*}{SL} &
  \multirow{-2}{*}{F} &
  D &
  \cellcolor[HTML]{F7D5D2}{\ul 0.797} &
  \cellcolor[HTML]{E5F5EE}92.5\% &
  \cellcolor[HTML]{F9DDDA}1.163 &
  \cellcolor[HTML]{EEF9F4}91.6\% &
  \cellcolor[HTML]{F7D3D0}{\ul 0.948} &
  \cellcolor[HTML]{E1F3EB}92.9\% &
  \cellcolor[HTML]{FBE6E4}0.846 &
  \cellcolor[HTML]{C7E8D8}97.0\% &
  \cellcolor[HTML]{F7D5D3}{\ul 0.939} &
  \cellcolor[HTML]{DBF1E6}93.5\% \\ \hline
 &
   &
   &
  M &
  \cellcolor[HTML]{FFFEFE}0.319 &
  \cellcolor[HTML]{CAEADA}95.1\% &
  \cellcolor[HTML]{FFFAFA}0.638 &
  \cellcolor[HTML]{F2FAF6}91.3\% &
  \cellcolor[HTML]{FFFBFB}0.490 &
  \cellcolor[HTML]{D1EDE0}94.4\% &
  \cellcolor[HTML]{F8D7D4}0.948 &
  \cellcolor[HTML]{C0E6D3}97.3\% &
  \cellcolor[HTML]{FEF7F7}0.599 &
  \cellcolor[HTML]{D0ECDF}94.5\% \\
\multirow{-2}{*}{TCM} &
  \multirow{-2}{*}{SSL} &
  \multirow{-2}{*}{F} &
  D &
  \cellcolor[HTML]{FFFFFF}\textbf{0.303} &
  \cellcolor[HTML]{D7EFE3}93.9\% &
  \cellcolor[HTML]{FFFFFF}\textbf{0.546} &
  \cellcolor[HTML]{EAF7F1}92.0\% &
  \cellcolor[HTML]{FFFFFF}\textbf{0.437} &
  \cellcolor[HTML]{DAF0E6}93.6\% &
  \cellcolor[HTML]{FDF0EF}{\ul 0.778} &
  \cellcolor[HTML]{C4E7D7}97.1\% &
  \cellcolor[HTML]{FFFFFF}\textbf{0.516} &
  \cellcolor[HTML]{D4EEE1}94.1\% \\ \hline
PointNet &
  SL &
  P &
  M &
  \cellcolor[HTML]{F5C6C2}0.973 &
  \cellcolor[HTML]{B9E3CF}96.7\% &
  \cellcolor[HTML]{F6CECA}1.428 &
  \cellcolor[HTML]{C5E8D7}95.6\% &
  \cellcolor[HTML]{F3BBB7}1.224 &
  \cellcolor[HTML]{B6E2CD}97.0\% &
  \cellcolor[HTML]{F5CAC6}1.032 &
  \cellcolor[HTML]{C5E8D7}97.1\% &
  \cellcolor[HTML]{F3BFBB}1.164 &
  \cellcolor[HTML]{BAE3D0}96.6\% \\ \hline
 &
   &
   &
  M &
  \cellcolor[HTML]{F5C7C3}0.970 &
  \cellcolor[HTML]{9BD7BA}98.4\% &
  \cellcolor[HTML]{FAE1DF}1.090 &
  \cellcolor[HTML]{71C69C}{\color[HTML]{333333} 99.4\%} &
  \cellcolor[HTML]{F6CBC8}1.040 &
  \cellcolor[HTML]{68C296}99.6\% &
  \cellcolor[HTML]{FAE1DF}0.880 &
  \cellcolor[HTML]{79C9A2}{\color[HTML]{333333} \textbf{99.2\%}} &
  \cellcolor[HTML]{F6D0CD}{\ul 0.995} &
  \cellcolor[HTML]{7BCAA4}\textbf{99.2\%} \\
 &
   &
   &
  D &
  \cellcolor[HTML]{F5CBC7}0.922 &
  \cellcolor[HTML]{75C79F}{\color[HTML]{333333} 99.3\%} &
  \cellcolor[HTML]{F8DAD7}1.216 &
  \cellcolor[HTML]{79C9A2}{\color[HTML]{333333} 99.2\%} &
  \cellcolor[HTML]{F2B6B1}1.282 &
  \cellcolor[HTML]{68C296}{\color[HTML]{333333} 99.6\%} &
  \cellcolor[HTML]{F6CECB}1.004 &
  \cellcolor[HTML]{B1E0C9}{\color[HTML]{333333} 97.8\%} &
  \cellcolor[HTML]{F4C5C1}1.106 &
  \cellcolor[HTML]{83CDA9}99.0\% \\
 &
   &
   &
  E &
  \cellcolor[HTML]{E98B83}1.669 &
  \cellcolor[HTML]{81CCA8}{\color[HTML]{333333} 99.0\%} &
  \cellcolor[HTML]{F5CBC7}1.483 &
  \cellcolor[HTML]{68C296}{\color[HTML]{333333} \textbf{99.6\%}} &
  \cellcolor[HTML]{EB958D}1.671 &
  \cellcolor[HTML]{64C193}99.7\% &
  \cellcolor[HTML]{FAE5E3}0.854 &
  \cellcolor[HTML]{92D3B4}{\color[HTML]{333333} 98.6\%} &
  \cellcolor[HTML]{EEA69F}1.419 &
  \cellcolor[HTML]{78C9A1}\textbf{99.2\%} \\
\multirow{-4}{*}{Ours} &
  \multirow{-4}{*}{SL} &
  \multirow{-4}{*}{P} &
  M+D+E &
  \cellcolor[HTML]{F5C7C3}0.968 &
  \cellcolor[HTML]{6CC499}{\color[HTML]{333333} \textbf{99.5\%}} &
  \cellcolor[HTML]{FAE3E1}{\ul 1.057} &
  \cellcolor[HTML]{6CC499}{\color[HTML]{333333} 99.5\%} &
  \cellcolor[HTML]{F5C9C5}1.065 &
  \cellcolor[HTML]{60BF90}\textbf{99.8\%} &
  \cellcolor[HTML]{FAE1DF}0.879 &
  \cellcolor[HTML]{BFE5D3}{\color[HTML]{333333} 97.3\%} &
  \cellcolor[HTML]{F6D0CD}0.992 &
  \cellcolor[HTML]{80CCA7}99.0\% \\ \hline
\end{tabular}%
}
\vspace{-7pt}
\caption{Quantitative results on MVSEC. The estimators are classified into model-based (MB), supervised learning (SL), self-supervised learning (SSL), frame-based (F), point-based (P). They are trained on MVSEC (M), DSEC (D), EVIMO-\texttt{imo} (E).}
\label{tab:mvsec_table}
\end{table*}

\begin{table*}[t]
\centering
\resizebox{0.95\textwidth}{!}{%
\begin{tabular}{lccccccccccccccc}
\hline
\multicolumn{2}{l}{} &
   &
   &
  \multicolumn{2}{c}{Scene\_13\_00} &
  \multicolumn{2}{c}{Scene\_13\_05} &
  \multicolumn{2}{c}{Scene\_14\_03} &
  \multicolumn{2}{c}{Scene\_14\_04} &
  \multicolumn{2}{c}{Scene\_14\_05} &
  \multicolumn{2}{c}{Average (8 scenes)} \\ \cline{5-16} 
\multicolumn{2}{l}{\multirow{-2}{*}{}} &
  \multirow{-2}{*}{Input} &
  \multirow{-2}{*}{\begin{tabular}[c]{@{}c@{}}Training\\ Set\end{tabular}} &
  PEE $\downarrow$ &
  \% Pos $\uparrow$ &
  PEE $\downarrow$ &
  \% Pos $\uparrow$ &
  PEE $\downarrow$ &
  \% Pos $\uparrow$ &
  PEE $\downarrow$ &
  \% Pos $\uparrow$ &
  PEE $\downarrow$ &
  \% Pos $\uparrow$ &
  PEE $\downarrow$ &
  \% Pos $\uparrow$ \\ \hline
MultiCM &
  MB &
  F &
  - &
  \cellcolor[HTML]{E8847B}1.509 &
  \cellcolor[HTML]{FFFFFF}53.2\% &
  \cellcolor[HTML]{E67C73}4.315 &
  \cellcolor[HTML]{FFFFFF}75.7\% &
  \cellcolor[HTML]{E67C73}1.611 &
  \cellcolor[HTML]{FFFFFF}79.2\% &
  \cellcolor[HTML]{E67C73}1.800 &
  \cellcolor[HTML]{FFFFFF}73.2\% &
  \cellcolor[HTML]{E67C73}2.768 &
  \cellcolor[HTML]{FFFFFF}72.9\% &
  \cellcolor[HTML]{E67C73}1.800 &
  \cellcolor[HTML]{FFFFFF}68.5\% \\
PCA &
  MB &
  P &
  - &
  \cellcolor[HTML]{E67C73}1.573 &
  \cellcolor[HTML]{FFFFFF}88.2\% &
  \cellcolor[HTML]{F4C6C2}2.035 &
  \cellcolor[HTML]{FFFFFF}87.5\% &
  \cellcolor[HTML]{E78077}1.580 &
  \cellcolor[HTML]{ECF7F2}91.9\% &
  \cellcolor[HTML]{E77E75}1.784 &
  \cellcolor[HTML]{FCFEFD}90.3\% &
  \cellcolor[HTML]{F1B5B0}1.823 &
  \cellcolor[HTML]{FFFFFF}89.4\% &
  \cellcolor[HTML]{E8857C}1.712 &
  \cellcolor[HTML]{FFFFFF}87.8\% \\ \hline
 &
   &
   &
  M &
  \cellcolor[HTML]{EB948D}1.370 &
  \cellcolor[HTML]{FFFFFF}71.9\% &
  \cellcolor[HTML]{F2BAB5}2.406 &
  \cellcolor[HTML]{FAFDFB}90.6\% &
  \cellcolor[HTML]{EC9A93}1.356 &
  \cellcolor[HTML]{FFFFFF}69.5\% &
  \cellcolor[HTML]{ED9E97}1.458 &
  \cellcolor[HTML]{FFFFFF}64.6\% &
  \cellcolor[HTML]{ED9F99}2.186 &
  \cellcolor[HTML]{FFFFFF}67.1\% &
  \cellcolor[HTML]{EC9B94}1.470 &
  \cellcolor[HTML]{FFFFFF}70.2\% \\
\multirow{-2}{*}{E-RAFT} &
  \multirow{-2}{*}{SL} &
  \multirow{-2}{*}{F} &
  D &
  \cellcolor[HTML]{F7D3D0}0.843 &
  \cellcolor[HTML]{FFFFFF}{\ul 88.9\%} &
  \cellcolor[HTML]{FAE1DF}1.185 &
  \cellcolor[HTML]{B0DFC9}97.5\% &
  \cellcolor[HTML]{FFFBFB}{\ul 0.517} &
  \cellcolor[HTML]{FFFFFF}88.1\% &
  \cellcolor[HTML]{FEF8F8}{\ul 0.538} &
  \cellcolor[HTML]{FFFFFF}85.9\% &
  \cellcolor[HTML]{FCEBEA}{\ul 0.908} &
  \cellcolor[HTML]{FFFFFF}86.3\% &
  \cellcolor[HTML]{FAE3E1}{\ul 0.705} &
  \cellcolor[HTML]{FFFFFF}87.9\% \\ \hline
 &
   &
   &
  M &
  \cellcolor[HTML]{F7D5D2}0.823 &
  \cellcolor[HTML]{FFFFFF}85.6\% &
  \cellcolor[HTML]{EDA09A}3.201 &
  \cellcolor[HTML]{C8E9D9}95.3\% &
  \cellcolor[HTML]{F2B6B1}1.111 &
  \cellcolor[HTML]{FFFFFF}86.3\% &
  \cellcolor[HTML]{EC9790}1.532 &
  \cellcolor[HTML]{FFFFFF}86.0\% &
  \cellcolor[HTML]{EA9088}2.445 &
  \cellcolor[HTML]{FFFFFF}82.2\% &
  \cellcolor[HTML]{EEA39D}1.383 &
  \cellcolor[HTML]{FFFFFF}84.6\% \\
\multirow{-2}{*}{TCM} &
  \multirow{-2}{*}{SSL} &
  \multirow{-2}{*}{F} &
  D &
  \cellcolor[HTML]{F9DBD8}{\ul 0.774} &
  \cellcolor[HTML]{FFFFFF}87.3\% &
  \cellcolor[HTML]{F1B5B0}2.541 &
  \cellcolor[HTML]{CAEADB}95.1\% &
  \cellcolor[HTML]{F7D2CF}0.872 &
  \cellcolor[HTML]{FFFFFF}87.8\% &
  \cellcolor[HTML]{F4C2BE}1.090 &
  \cellcolor[HTML]{FFFFFF}86.5\% &
  \cellcolor[HTML]{F3C0BB}1.640 &
  \cellcolor[HTML]{FFFFFF}84.1\% &
  \cellcolor[HTML]{F3BDB9}1.105 &
  \cellcolor[HTML]{FFFFFF}85.3\% \\ \hline
PointNet &
  SL &
  P &
  E &
  \cellcolor[HTML]{F2BBB6}1.047 &
  \cellcolor[HTML]{FFFFFF}88.1\% &
  \cellcolor[HTML]{FBE9E8}{\ul 0.924} &
  \cellcolor[HTML]{AFDFC8}{\ul 97.7\%} &
  \cellcolor[HTML]{F7D5D2}0.848 &
  \cellcolor[HTML]{9FD8BD}{\ul 98.3\%} &
  \cellcolor[HTML]{F8D6D3}0.892 &
  \cellcolor[HTML]{BEE5D3}{\ul 96.2\%} &
  \cellcolor[HTML]{FAE3E1}1.053 &
  \cellcolor[HTML]{BAE3D0}{\ul 96.6\%} &
  \cellcolor[HTML]{F6CDCA}0.933 &
  \cellcolor[HTML]{CCEADC}{\ul 95.0\%} \\ \hline
 &
   &
   &
  M &
  \cellcolor[HTML]{FAE2E0}0.713 &
  \cellcolor[HTML]{C5E8D7}95.6\% &
  \cellcolor[HTML]{FFFEFE}0.269 &
  \cellcolor[HTML]{75C79F}99.3\% &
  \cellcolor[HTML]{FBE9E7}0.676 &
  \cellcolor[HTML]{8AD0AE}98.8\% &
  \cellcolor[HTML]{FCEDEC}0.651 &
  \cellcolor[HTML]{A7DCC3}98.1\% &
  \cellcolor[HTML]{FDF1F0}0.806 &
  \cellcolor[HTML]{A3DAC0}98.2\% &
  \cellcolor[HTML]{FDF1F0}0.551 &
  \cellcolor[HTML]{B5E1CC}97.1\% \\
 &
   &
   &
  D &
  \cellcolor[HTML]{FDF1F0}0.590 &
  \cellcolor[HTML]{BAE3D0}96.6\% &
  \cellcolor[HTML]{FFFFFF}\textbf{0.230} &
  \cellcolor[HTML]{60BF90}\textbf{99.8\%} &
  \cellcolor[HTML]{FDF4F4}0.575 &
  \cellcolor[HTML]{60BF90}\textbf{99.8\%} &
  \cellcolor[HTML]{FDF0EF}0.625 &
  \cellcolor[HTML]{6CC499}{\ul \textbf{99.5\%}} &
  \cellcolor[HTML]{FFFFFF}\textbf{0.567} &
  \cellcolor[HTML]{71C69C}\textbf{99.4\%} &
  \cellcolor[HTML]{FEF9F9}0.463 &
  \cellcolor[HTML]{ACDEC6}\textbf{97.9\%} \\
 &
   &
   &
  E &
  \cellcolor[HTML]{FFFCFB}0.497 &
  \cellcolor[HTML]{B9E3CF}\textbf{96.7\%} &
  \cellcolor[HTML]{FEFAFA}0.399 &
  \cellcolor[HTML]{79C9A2}99.2\% &
  \cellcolor[HTML]{FFFFFF}\textbf{0.478} &
  \cellcolor[HTML]{79C9A2}99.2\% &
  \cellcolor[HTML]{FFFBFA}0.515 &
  \cellcolor[HTML]{8AD0AE}98.8\% &
  \cellcolor[HTML]{FFFFFE}0.584 &
  \cellcolor[HTML]{92D3B4}98.6\% &
  \cellcolor[HTML]{FFFDFD}0.423 &
  \cellcolor[HTML]{ADDEC6}97.9\% \\
\multirow{-4}{*}{Ours} &
  \multirow{-4}{*}{SL} &
  \multirow{-4}{*}{P} &
  M+D+E &
  \cellcolor[HTML]{FFFFFF}\textbf{0.465} &
  \cellcolor[HTML]{BEE5D3}96.2\% &
  \cellcolor[HTML]{FFFDFD}0.308 &
  \cellcolor[HTML]{79C9A2}99.2\% &
  \cellcolor[HTML]{FEF8F7}0.544 &
  \cellcolor[HTML]{75C79F}99.3\% &
  \cellcolor[HTML]{FFFFFF}\textbf{0.467} &
  \cellcolor[HTML]{8AD0AE}98.8\% &
  \cellcolor[HTML]{FFFFFF}0.568 &
  \cellcolor[HTML]{96D5B7}98.5\% &
  \cellcolor[HTML]{FFFFFF}\textbf{0.396} &
  \cellcolor[HTML]{ADDEC7}97.8\% \\ \hline
\end{tabular}%
}
\vspace{-7pt}
\caption{Quantitative results on the first five scenes of EVIMO2-\texttt{imo}'s evaluation set. The complete table is in Appendix \ref{app:evimo_per_scene}.}
\label{tab:evimo_table}
\vspace{-7pt}
\end{table*}

As shown by the quantitative results in Table \ref{tab:mvsec_table},
\noindent\textbf{our point-based method is on par with state-of-the-art frame-based methods despite using only local information.} While TCM achieves lower PEE than our method on MVSEC, this is because MVSEC mainly features slow-moving, static scenes without independently moving objects. In these scenarios, TCM and E-RAFT using CNNs, can leverage larger receptive fields (through stacked convolution and pooling layers) to smooth the flow predictions. 
As we will show in Section \ref{sec:s42_EVIMO}, TCM and E-RAFT's performance  deteriorates when scenes include independently moving objects, whereas our method remains robust.

\noindent\textbf{Our method outperforms the point-based estimators, PCA (MB) and PointNet (SL)} by achieving lower PEE and higher $\% Pos$, highlighting the benefits of supervised training. Additionally, PointNet requires explicit grouping of events into neighborhoods and sampling, while our method does not. This allows us to use information from more neighboring events to predict the flows and yield better performance.
In cross-dataset evaluations, our estimator outperforms PCA and PointNet when trained on DSEC and evaluated on MVSEC even though PointNet was trained on MVSEC. When our method is trained on EVIMO-\texttt{imo}, the performance remains comparable though the EVIMO-\texttt{imo} train set containing only 2.74 minutes of data.

\subsection{Evaluation on EVIMO2}
\label{sec:s42_EVIMO}

\begin{figure*}[t]
    \centering
    \includegraphics[width=\linewidth]{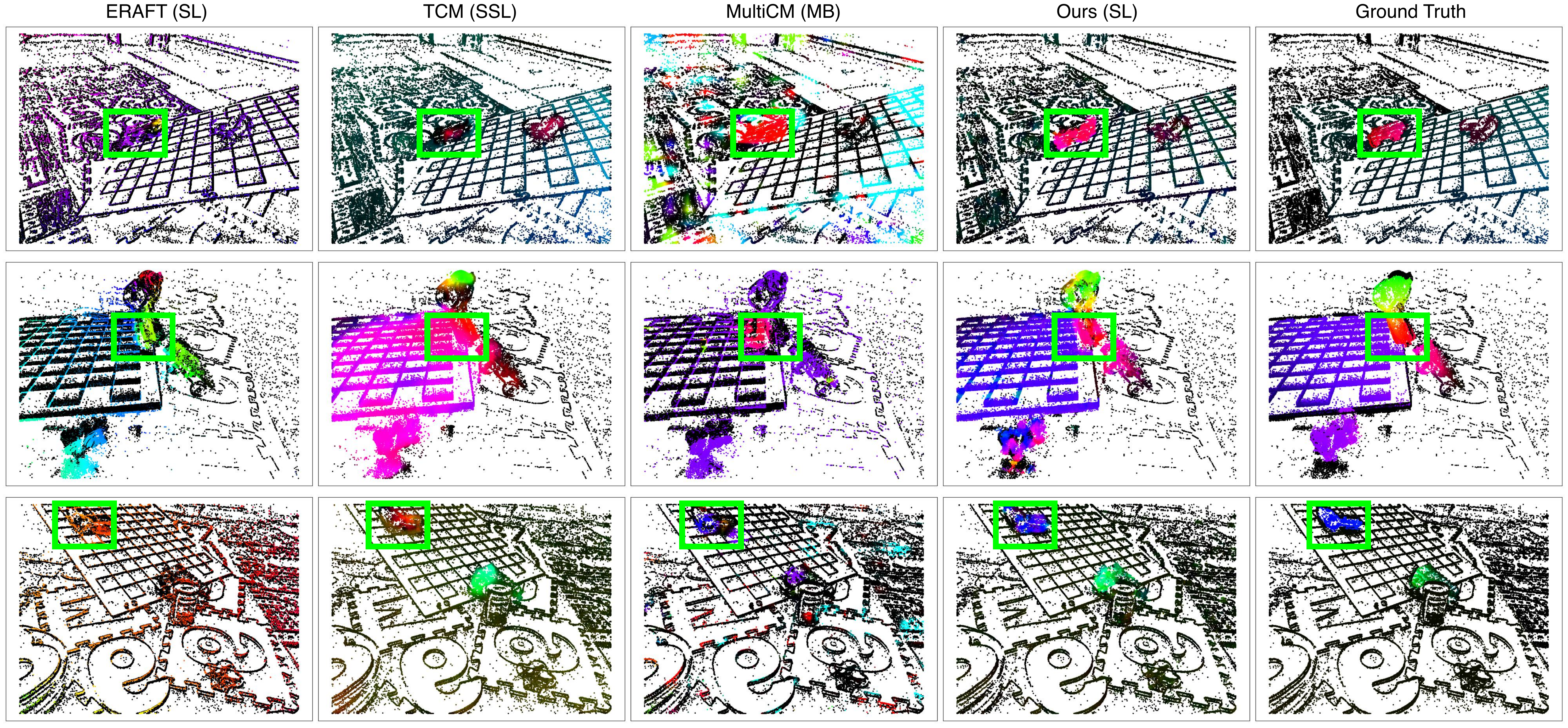}
    \vspace{-18pt}
    \caption{Visualization of flow prediction on EVIMO2-\texttt{imo}. The flows are displayed in HSV color space. Our method effectively preserves the boundary of the IMOs, while other methods fail.}
    \label{fig:EVIMO_vis}
    \vspace{-7pt}
\end{figure*}

EVIMO2-\texttt{imo} sequences feature fast independently moving objects. Table \ref{tab:evimo_table} presents quantitative results on the samsung\_mono camera and Figure \ref{fig:EVIMO_vis} visualizes the flow prediction colored in HSV. As shown on the Table \ref{tab:evimo_table}, \textbf{our method, significantly outperforms all compared estimators in the presence of independently moving objects even when trained on other datasets.}
In addition, as shown in Figure \ref{fig:EVIMO_vis}, our method effectively preserves the boundaries of the IMOs, closely matching the ground truth.

\textbf{Our estimator shows highly consistent performance and strong generalizability across different scenes, motion types and datasets} as shown in  Table \ref{tab:evimo_table}. 
This is due to the model's use of geometric features from local event neighborhoods, which capture domain-invariant patterns essential for normal flow estimation. Besides, our approach employs normalized camera coordinates, utilizes extensive data augmentation, and prevents global information from affecting estimates in regions with independently moving objects (IMOs), which all lead to the strong performance.


\subsection{Evaluation on DSEC}
\label{sec:s43_DSEC}

Since DSEC does not provide GT optical flow, we cannot compute error metrics for normal flow evaluation.
We only provide a  qualitative evaluation. The flow prediction videos in \href{https://drive.google.com/drive/u/3/folders/1gkmUyZX5VRf8DxiBKL9CSdWdifjqZVq3}{this drive} show \textbf{our method performs equally well even when trained on different datasets}. Our method produces consistently accurate predictions though the event density of MVSEC and the motion type of EVIMO2 are very different from those in DSEC, 

\section{Ablation Studies}
\vspace{-3pt}
We perform extensive ablation studies on our proposed method. We summarize the conclusion in this section, leaving the experiment statistics to Appendix \ref{app:ablation_studies}.

\noindent\textbf{Effectiveness of motion field loss.} We compare the model's performance against a baseline trained with conventional optical flow loss function. The model trained with our proposed motion field loss achieves significantly lower error, highlighting the effectiveness of our loss function.

\noindent\textbf{Effectiveness of uncertainty quantification (UQ).} We find the flow prediction errors are positively correlated with the uncertainty scores, highlighting the effectiveness of the UQ. We also study how many ensembles of predictions are needed to generate reliable UQ.

\noindent\textbf{Runtime.} Our estimator takes about three seconds to predict 80,000 per-event flows on an RTX A5000 GPU with 24 GB of memory, with memory usage below 6 GB. We also include statistics about event density of each dataset.

\begin{figure}[t]
    \centering
    \includegraphics[width=0.93\linewidth]{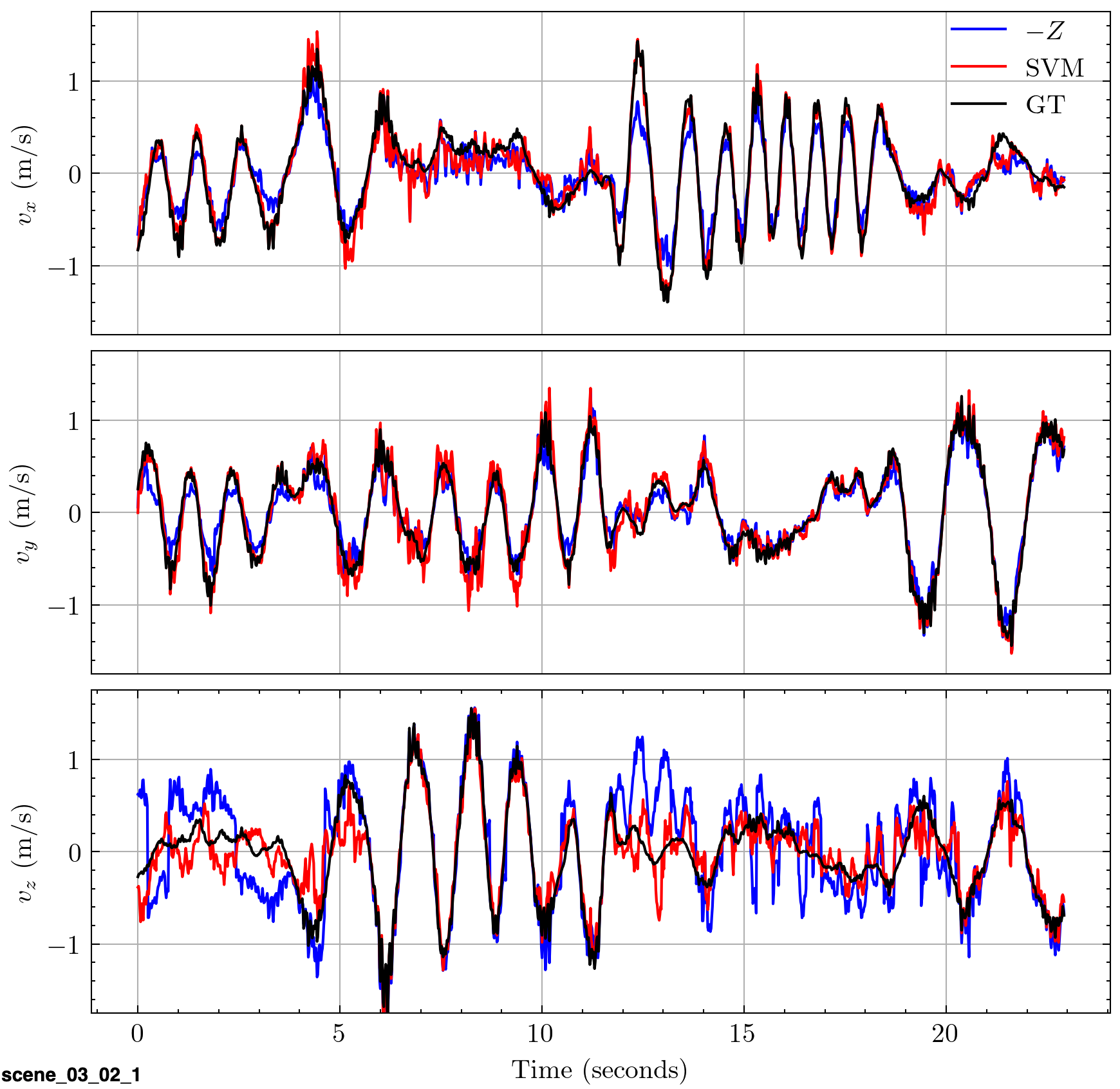}
    \caption{Comparison of estimated translation velocity produced by maximizing negative depth \cite{barranco2021joint} and our SVM-based method. Our approach yields more accurate predictions.}
    \label{fig:egomotion}
\end{figure}

\begin{table}[ht]
  \centering
  \resizebox{\columnwidth}{!}{
  \begin{tabular}{@{}lccccc@{}}
    \toprule
  \texttt{Seq.} &  \texttt{sfm 03\_04\_0} & \texttt{sfm 03\_02\_0} & \texttt{sfm 03\_02\_1} & \texttt{sfm 03\_00\_0} & \texttt{sfm 03\_03\_0} \\
    \midrule
    Method & \multicolumn{5}{c}{RMS V (m/s)} \\
    \midrule
Depth Pos. & 0.217 & 0.500 & 0.243 & 0.237 & 0.135 \\
SVM & \textbf{0.136} & \textbf{0.227} & \textbf{0.156} & \textbf{0.141} & \textbf{0.116} \\
    \midrule
    \midrule
  \texttt{Seq.} &  \texttt{sfm 03\_01\_0} & \texttt{sfm 03\_02\_2} & \texttt{sfm 03\_03\_2} & \texttt{sfm 03\_02\_3} & \texttt{sfm 03\_03\_1}\\
    \midrule
    Method &  \multicolumn{5}{c}{RMS V (m/s)} \\
    \midrule
Depth Pos. & 0.369 & \textbf{0.091} & 0.137 & \textbf{0.213} & 0.265 \\
SVM & \textbf{0.219} & 0.127 & \textbf{0.108} & 0.279 & \textbf{0.136} \\
    \bottomrule
  \end{tabular}}
  \caption{Egomotion estimation error on EVIMO2-\texttt{sfm}.}
  \label{tab:egomotion}
  \vspace{-21pt}
\end{table}

\vspace{-3pt}
\section{Experiments on Egomotion Estimation}
\vspace{-3pt}
\label{sec:s50_egomotion}
We evaluate our egomotion solver on the EVIMO2 \texttt{sfm} split, which includes scenes with fast camera motions and a focus of expansion far from the camera frame. The normal flow is estimated using our method trained on the EVIMO2-\texttt{imo} training split. Table \ref{tab:egomotion} and Figure \ref{fig:egomotion} compares our solver using SVM, to a method based on \cite{barranco2021joint} that estimates translation by maximizing negative depth using rotation estimates from the IMU. The results demonstrate that our normal flow predictions and egomotion solver yield more accurate egomotion estimations. In Figure \ref{fig:egomotion} the estimated translation direction is scaled to m/s using ground truth. Plots of the predictions for all scenes are provided in Appendix \ref{app:evimo_egomotion}.

\section{Conclusion and Future Work}
We introduce a point-based method for normal flow estimation that overcomes limitations in existing model and learning based methods by using local information without explicit grouping. This allows improved cross domain transfer and good performance in the presence of independently moving objects. While efficient, the method may face computational challenges with higher camera resolutions. Thus, future work can consider optimizing the encoding and transformation to normal flow. Additionally, our method currently depends on ground-truth flow for training. A self-supervised approach could further improve its usability. Finally,
while our method benefits from using local information, future work can consider careful incorporation of global information that may be beneficial for tasks such as action recognition and object detection.


\newpage
{
    \small
    \bibliographystyle{ieeenat_fullname}
    \bibliography{main}
}


\clearpage
\setcounter{page}{1}
\maketitlesupplementary

\begin{figure*}[t]
    \centering
    \includegraphics[width=0.95\linewidth]{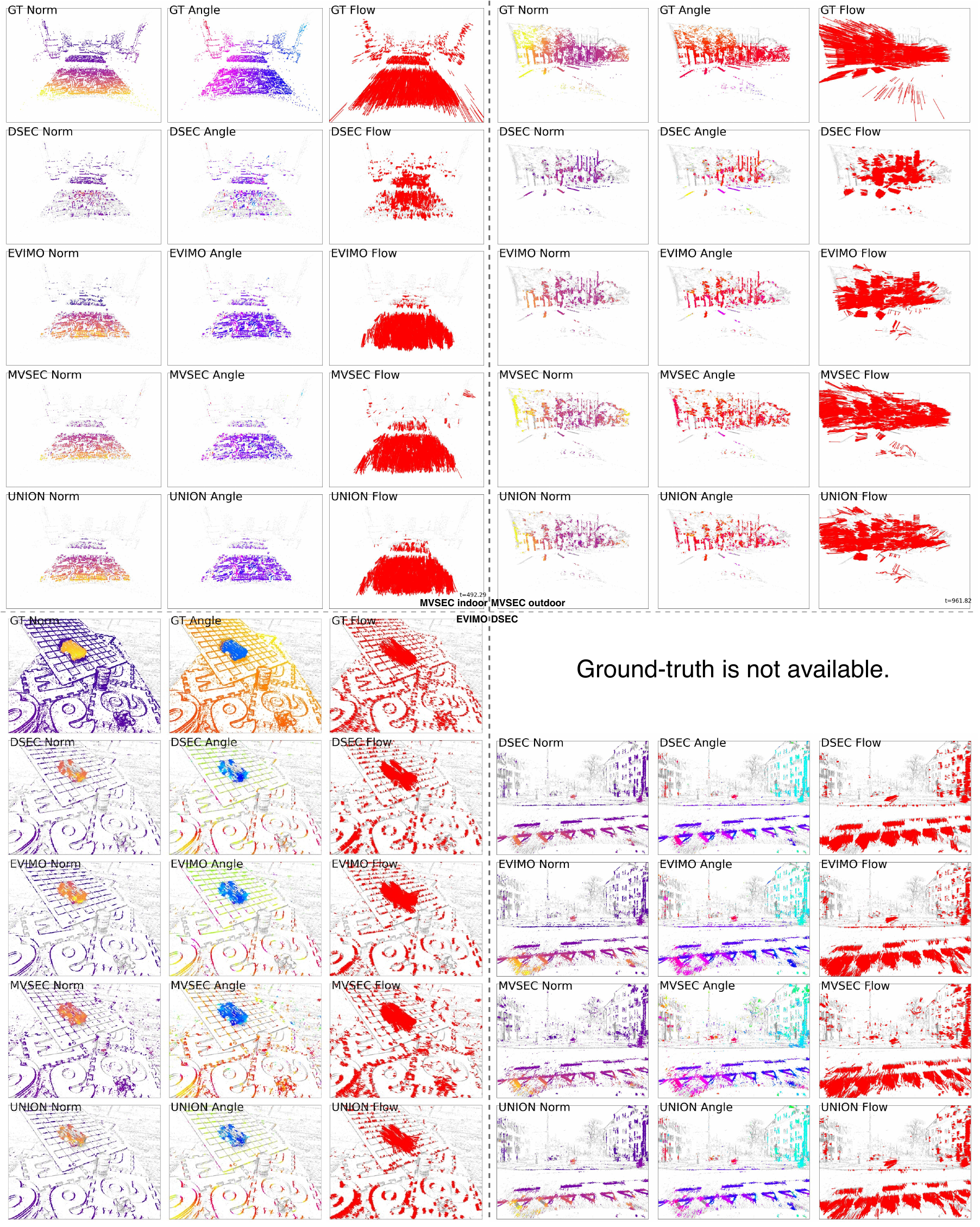}
    \caption{Screenshot of the flow prediction videos. Each row displays the norm, angle, and flow fields of both ground-truth and predicted flows. The first row visualizes the ground-truth optical flow, while subsequent rows show model predictions trained on each dataset. To illustrate the flow field, we sample 5,000 flow points for visualization. If a pixel is gray, it means the flow prediction has a high uncertainty.}
    \label{fig:prediction_videos}
\end{figure*}

\newpage

\section{List of Flow Prediction Videos}
\label{app:flow_prediction_video}

We provide flow prediction videos for each evaluated dataset—MVSEC, EVIMO2, and DSEC. These visualizations showcase predictions from models trained on each of the three datasets. Figure \ref{fig:prediction_videos} shows some screenshots of the video. The videos are in \href{https://drive.google.com/drive/u/3/folders/1gkmUyZX5VRf8DxiBKL9CSdWdifjqZVq3}{this link}. The enumeration of videos are as followed:
\begin{itemize}
	\item DSEC\_eval\_interlaken\_00\_a.mov
	\item DSEC\_eval\_interlaken\_00\_b.mov
	\item DSEC\_eval\_interlaken\_01\_a.mov
	\item DSEC\_eval\_thun\_01\_a.mov
	\item DSEC\_eval\_thun\_01\_b.mov
	\item DSEC\_eval\_zurich\_city\_12\_a.mov
	\item DSEC\_eval\_zurich\_city\_13\_a.mov
	\item DSEC\_eval\_zurich\_city\_13\_b.mov
	\item DSEC\_eval\_zurich\_city\_14\_a.mov
	\item DSEC\_eval\_zurich\_city\_14\_b.mov
	\item DSEC\_eval\_zurich\_city\_14\_c.mov
	\item DSEC\_eval\_zurich\_city\_15\_a.mov
	\item EVIMO\_eval\_scene13\_dyn\_test\_00\_000000.mov
	\item EVIMO\_eval\_scene13\_dyn\_test\_05\_000000.mov
	\item EVIMO\_eval\_scene14\_dyn\_test\_03\_000000.mov
	\item EVIMO\_eval\_scene14\_dyn\_test\_04\_000000.mov
	\item EVIMO\_eval\_scene14\_dyn\_test\_05\_000000.mov
	\item EVIMO\_eval\_scene15\_dyn\_test\_01\_000000.mov
	\item EVIMO\_eval\_scene15\_dyn\_test\_02\_000000.mov
	\item EVIMO\_eval\_scene15\_dyn\_test\_05\_000000.mov
    \item EVIMO\_sfm\_scene\_03\_00\_000000.mov
    \item EVIMO\_sfm\_scene\_03\_01\_000000.mov
    \item EVIMO\_sfm\_scene\_03\_02\_000000.mov
    \item EVIMO\_sfm\_scene\_03\_02\_000001.mov
    \item EVIMO\_sfm\_scene\_03\_02\_000002.mov
    \item EVIMO\_sfm\_scene\_03\_02\_000003.mov
    \item EVIMO\_sfm\_scene\_03\_03\_000000.mov
    \item EVIMO\_sfm\_scene\_03\_03\_000001.mov
    \item EVIMO\_sfm\_scene\_03\_03\_000002.mov
    \item EVIMO\_sfm\_scene\_03\_04\_000000.mov
	\item MVSEC\_eval\_indoor\_flying1.mov
	\item MVSEC\_eval\_indoor\_flying2.mov
	\item MVSEC\_eval\_indoor\_flying3.mov
	\item MVSEC\_eval\_outdoor\_day1.mov
\end{itemize}

\begin{table*}[t]
\centering
\resizebox{\textwidth}{!}{%
\begin{tabular}{@{}lcccccccccccccccccc@{}}
\toprule
 &
  \multicolumn{2}{c}{13\_00} &
  \multicolumn{2}{c}{13\_05} &
  \multicolumn{2}{c}{14\_03} &
  \multicolumn{2}{c}{14\_04} &
  \multicolumn{2}{c}{14\_05} &
  \multicolumn{2}{c}{15\_01} &
  \multicolumn{2}{c}{15\_02} &
  \multicolumn{2}{c}{15\_05} &
  \multicolumn{2}{c}{Average} \\ \cmidrule(l){2-19} 
 &
  PEE &
  \%Pos &
  PEE &
  \%Pos &
  PEE &
  \%Pos &
  PEE &
  \%Pos &
  PEE &
  \%Pos &
  PEE &
  \%Pos &
  PEE &
  \%Pos &
  PEE &
  \%Pos &
  PEE &
  \%Pos \\ \midrule
Norm + Direction Loss &
  0.972 &
  90.0\% &
  0.912 &
  98.1\% &
  0.749 &
  98.9\% &
  0.762 &
  96.9\% &
  1.150 &
  97.3\% &
  0.559 &
  96.9\% &
  0.610 &
  94.5\% &
  1.274 &
  91.5\% &
  0.87 &
  95.5\% \\
Ours &
  0.497 &
  96.7\% &
  0.399 &
  99.2\% &
  0.478 &
  99.2\% &
  0.515 &
  98.8\% &
  0.584 &
  98.6\% &
  0.286 &
  98.1\% &
  0.274 &
  96.8\% &
  0.354 &
  95.5\% &
  0.42 &
  97.9\% \\ \midrule
Difference &
  {\color[HTML]{009901} $\downarrow$ 0.475} &
  {\color[HTML]{009901} $\uparrow$ 6.7\%} &
  {\color[HTML]{009901} $\downarrow$ 0.513} &
  {\color[HTML]{009901} $\uparrow$ 1.1\%} &
  {\color[HTML]{009901} $\downarrow$ 0.271} &
  {\color[HTML]{009901} $\uparrow$ 0.3\%} &
  {\color[HTML]{009901} $\downarrow$ 0.247} &
  {\color[HTML]{009901} $\uparrow$ 1.9\%} &
  {\color[HTML]{009901} $\downarrow$ 0.566} &
  {\color[HTML]{009901} $\uparrow$ 1.3\%} &
  {\color[HTML]{009901} $\downarrow$ 0.273} &
  {\color[HTML]{009901} $\uparrow$ 1.2\%} &
  {\color[HTML]{009901} $\downarrow$ 0.336} &
  {\color[HTML]{009901} $\uparrow$ 2.3\%} &
  {\color[HTML]{009901} $\downarrow$ 0.92} &
  {\color[HTML]{009901} $\uparrow$ 4.0\%} &
  {\color[HTML]{009901} $\downarrow$ 0.45} &
  {\color[HTML]{009901} $\uparrow$ 2.4\%} \\ \bottomrule
\end{tabular}%
}
\caption{Comparison between the estimator trained with our motion field loss function and the one trained with the standard norm-plus-direction loss function. Using our motion field function significantly improves the model's performance.}
\label{tab:motion_field_ablation}
\end{table*}

\begin{figure*}[t]
    \centering
    \includegraphics[width=\linewidth]{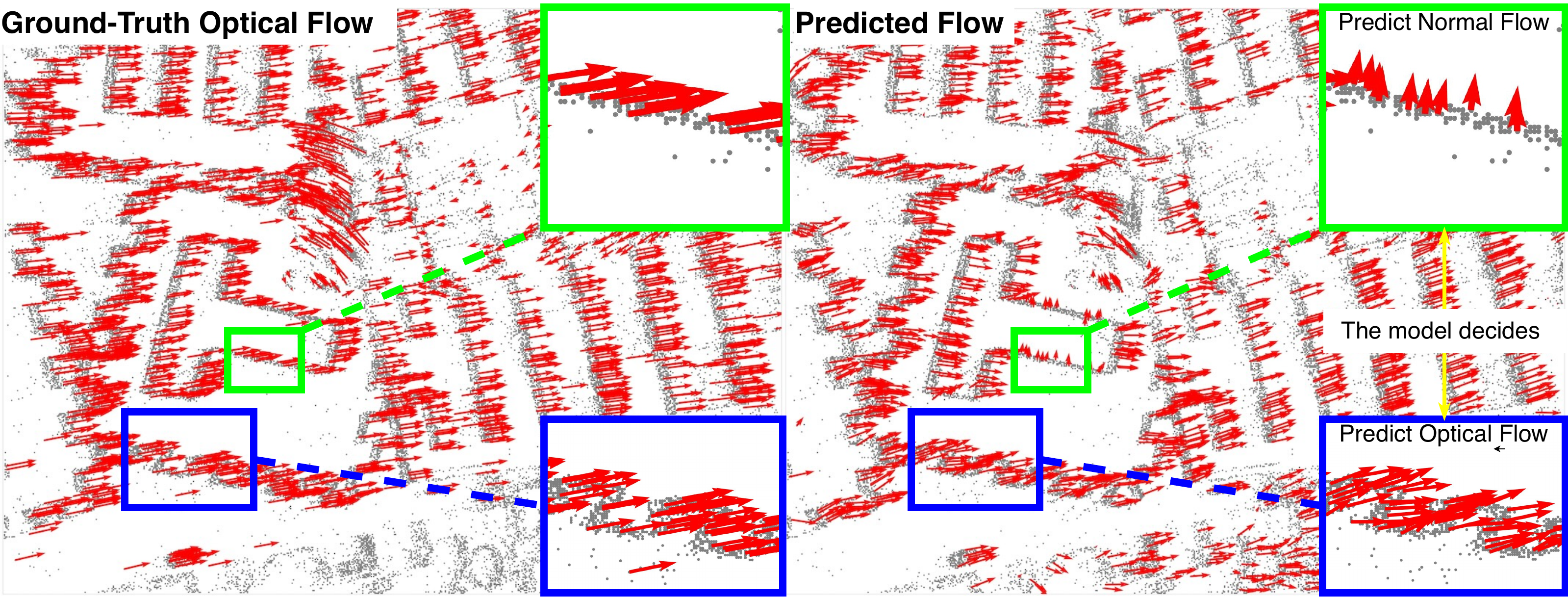}
    \caption{The model can choose between estimating the full optical flow or normal flow depending on the texture of the local region. If the local texture is rich enough (e.g. a corner), the model will estimate full optical flow. If the local texture only contains strong edges, the model will estimate normal flow.}
    \label{fig:effect_motion_field}
    \vspace{-7pt}
\end{figure*}


\newpage
\section{Dataset Preprocessing}
\label{app:dataset_preprocess}
In this section, we detail how to preprocess the data to obtain undistorted normalized per-event optical flow on MVSEC, EVIMO2, and DSEC. 

\noindent\textbf{MVSEC \& EVIMO2} both provide frame-based forward optical flows in the distorted camera coordinates. We first interpolate the flow in the time domain. If an event $(t, x, y)$ lies between $t_0$ and $t_1$, the optical flow at this event is computed as:
\[\bu(t, x, y)=\frac{t-t_0}{t_1-t_0}flow(t_1, x, y) + \frac{t_1-t}{t_1-t_0}flow(t_0, x, y)\]
After this, we convert the per-event distorted flow in the raw pixel coordinates into undistorted flow in the normalized pixel coordinates using \texttt{cv2.undistortPoints}.
\begin{align*}
    \mathrm{start} &= \texttt{cv2.undistortPoints}(x, y, K, D) \\
    \mathrm{end} &= \texttt{cv2.undistortPoints}(x+\bu_x, y+\bu_y, K, D) \\
    \mathrm{out} &= (\mathrm{end} - \mathrm{start}) / (t_1 - t_0)
\end{align*}
This will transform the flow into undistorted normalized camera coordinates, with unit normalized pixel per second.

\noindent\textbf{DSEC}, different from the previous two datasets, provides frame-based forward optical flow and backward optical flow, which can be used to obtain more accurate per-event optical flow. Specifically, we let
\begin{align*}
  &flow(t_1, x, y) = \frac{1}{2}(\\
  &\qquad flow\_forward(t_1, x, y) - flow\_backward(t_1, x, y) \\
  &)
\end{align*}
The following procedures are the same as the previous two datasets.

\section{Implementation Details of Our Model}
\label{app:our_model}
We transform the event pixels and flows into undistorted, normalized camera coordinates as explained in Appendix \ref{app:dataset_preprocess}. The resulting flows are then scaled such that their unit is in pixels per second. After this scaling, the flow norms fall within a range of 0 to 3. 

During training, we randomly sample an event, using a uniform distribution over the logarithm of the flow norm, within the range of 0.01 to 3. We then slice the event stream around the sampled event to create the training samples. We apply the data augmentation techniques described in Section \ref{sec:data_augmentation}. The pixel radius parameters ($\delta x$, $\delta y$ in Eqn. \eqref{eqn:local_events}) are set to 0.02, which correspond to 4.5, 10.4, and 11.1 pixels for the MVSEC, EVIMO2, and DSEC datasets, respectively, measured in terms of raw pixels. The time radius ($\delta t$ in Eqn. \eqref{eqn:local_events}) is 20 ms. The parameter $\epsilon$ in Eqn. \eqref{eqn:radial} is set to 0.1. The dimension of the local event encoding is 384. We remove the predictions with circular standard deviation larger than 0.3 (Section \ref{sec:uncertainty_quantification}). If the events size within 20 ms is larger than 80,000, we randomly sample 80,000 events from the 20 ms interval.

\section{Ablation Studies}
\label{app:ablation_studies}

We use the EVIMO2-\texttt{imo} dataset for our ablation studies because it presents challenging scenarios with independently moving objects. The models are trained on EVIMO2-\texttt{imo} training set to better capture the impact of the ablated factors.

\subsection{Effect of Motion Field Loss}
\label{app:motion_field_loss}

We show that our estimator benefits from being trained on the novel motion field loss by comparing to an optical flow loss. While variations on average end-point error (AEE) are typically used for supervised training of optical flow estimators, our method, which is designed to estimate normal flow, does not converge when trained with such losses. Thus, we designed the following norm + direction loss to train our estimator to estimate optical flow, defined as follows:
\begin{align*}
    \mathcal{L}_1 &= \log\Big(\frac{\epsilon + ||\bu||}{\epsilon + ||\hat{\bu}||}\Big)^2 \\
    \mathcal{L}_2 &= -\frac{\bu \cdot \hat{\bu}}{||\bu|| \cdot ||\hat{\bu}||} \\
    \mathcal{L} &= \mathcal{L}_1 + \mathcal{L}_2
\end{align*}
Where $\hat{\bu}$ is the output of our method when being trained with this optical flow loss.

As shown in Table \ref{tab:motion_field_ablation}, our motion field loss function significantly enhances the estimator's performance in terms of PEE and $\% Pos$.

In addition, we analyze the behavior of our estimator qualitatively in Figure \ref{fig:effect_motion_field}. After the model is trained using our motion field loss function, the model can choose between estimating full optical flow or normal flow depending on the texture of the local regions. This further justifies the effect of our motion field loss function.

\subsection{Effect of Uncertainty Quantification}
\label{app:uncertainty_quantification}
Table \ref{tab:ensemble_ablation} provides a comprehensive analysis of the estimator's performance, showing prediction errors alongside the percentage of confident predictions across various ensemble sizes and uncertainty thresholds. The positive correlation observed between prediction errors and uncertainty scores underscores the effectiveness of the uncertainty quantification. Our results indicate that an uncertainty threshold between 0.3 and 0.6 achieves an optimal balance between valid prediction rates and accuracy. Additionally, the table reveals that 3 to 4 ensemble predictions are sufficient for consistent uncertainty estimation, though larger ensembles generally yield improved performance. For scenarios where runtime is not a constraint, employing larger ensembles can enhance prediction accuracy.

\subsection{Runtime and Memory Usage}
\label{app:runtime}
Table \ref{tab:computation_cost} reports the computational cost of our estimator. Table \ref{tab:event_density} reports the statistics about the event density of some selected scenes from each dataset.

\section{Per-Scene Normal Flow Evaluation on EVIMO2}
\label{app:evimo_per_scene}
We present the per-scene normal flow evaluation on EVIMO2-\texttt{imo}, as shown in Table \ref{tab:evimo_per_scene_table}.

\section{Per-Scene Egomotion Evaluation on EVIMO2}
\label{app:evimo_egomotion}
\begin{figure*}
    \centering
    \includegraphics[width=\linewidth]{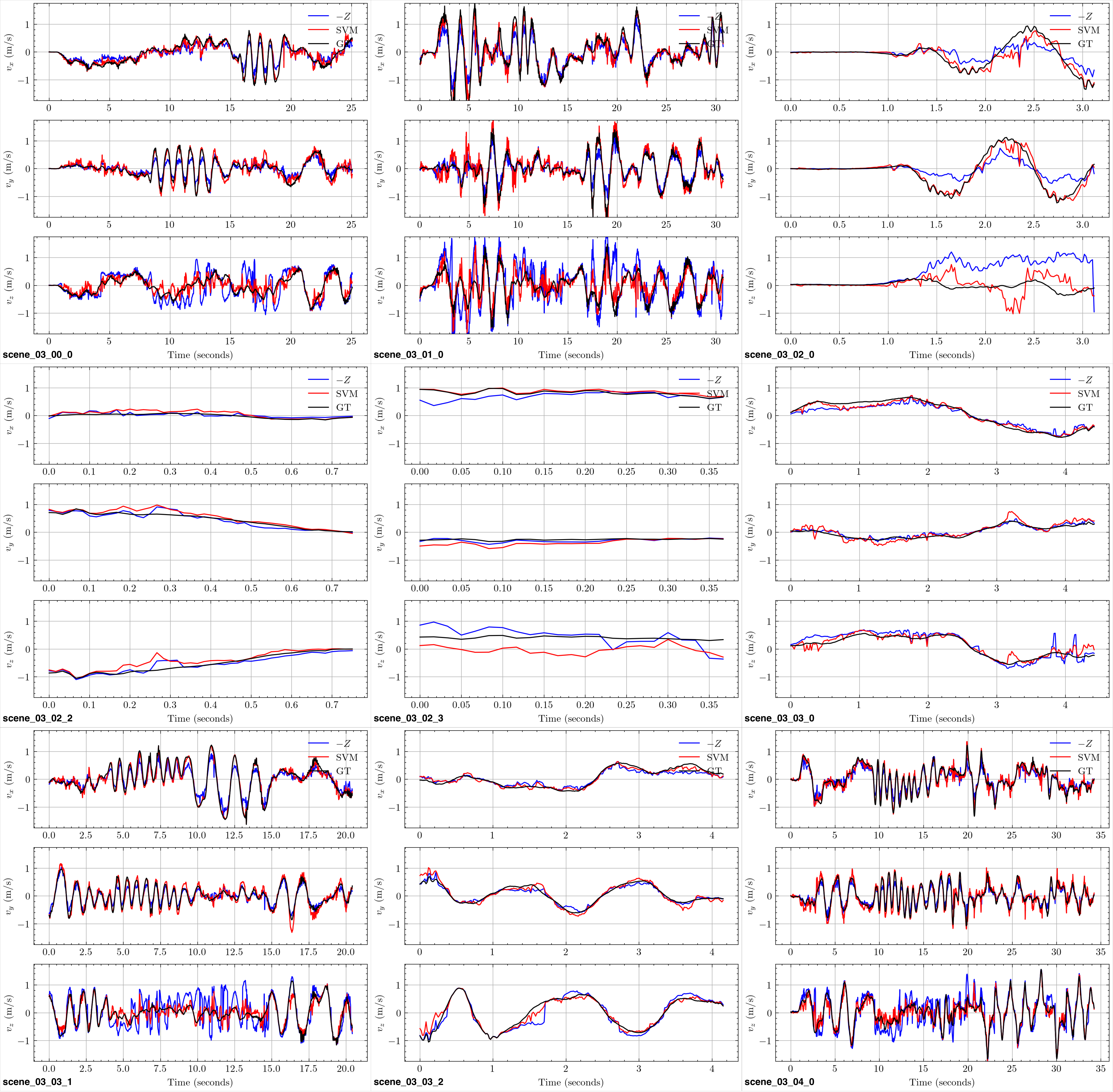}
    \caption{Per-scene egomotion evaluation on EVIMO2 \texttt{sfm} split.}
    \label{fig:per-scene-egomotion}
\end{figure*}

We present the per-scene egomotion evaluation on EVIMO2, as shown in Figure \ref{fig:per-scene-egomotion}.

\newpage

\begin{table}[t]
\centering
\resizebox{\columnwidth}{!}{%
\begin{tabular}{lcccc}
\multicolumn{1}{c|}{\textbf{\% Pos}} &
  \cellcolor[HTML]{FFFFFF}num\_ensemble=2 &
  num\_ensemble=4 &
  num\_ensemble=6 &
  num\_ensemble=10 \\ \hline
\multicolumn{1}{l|}{conf\_thres=0.1} &
  \cellcolor[HTML]{79C9A2}97.2\% &
  \cellcolor[HTML]{6BC398}97.7\% &
  \cellcolor[HTML]{57BB8A}98.4\% &
  \cellcolor[HTML]{5DBE8E}98.2\% \\
\multicolumn{1}{l|}{conf\_thres=0.2} &
  \cellcolor[HTML]{81CCA7}96.9\% &
  \cellcolor[HTML]{63C092}98.0\% &
  \cellcolor[HTML]{5ABD8C}98.3\% &
  \cellcolor[HTML]{57BB8A}98.4\% \\
\multicolumn{1}{l|}{conf\_thres=0.3} &
  \cellcolor[HTML]{8CD1AF}96.5\% &
  \cellcolor[HTML]{6BC398}97.7\% &
  \cellcolor[HTML]{63C092}98.0\% &
  \cellcolor[HTML]{60BF90}98.1\% \\
\multicolumn{1}{l|}{conf\_thres=0.4} &
  \cellcolor[HTML]{94D4B5}96.2\% &
  \cellcolor[HTML]{73C79E}97.4\% &
  \cellcolor[HTML]{6BC398}97.7\% &
  \cellcolor[HTML]{65C194}97.9\% \\
\multicolumn{1}{l|}{conf\_thres=0.5} &
  \cellcolor[HTML]{9CD7BA}95.9\% &
  \cellcolor[HTML]{79C9A2}97.2\% &
  \cellcolor[HTML]{70C69C}97.5\% &
  \cellcolor[HTML]{6BC398}97.7\% \\
\multicolumn{1}{l|}{conf\_thres=0.6} &
  \cellcolor[HTML]{A2DABE}95.7\% &
  \cellcolor[HTML]{81CCA7}96.9\% &
  \cellcolor[HTML]{76C8A0}97.3\% &
  \cellcolor[HTML]{70C69C}97.5\% \\
\multicolumn{1}{l|}{conf\_thres=0.7} &
  \cellcolor[HTML]{A7DCC2}95.5\% &
  \cellcolor[HTML]{86CEAB}96.7\% &
  \cellcolor[HTML]{7BCAA3}97.1\% &
  \cellcolor[HTML]{79C9A2}97.2\% \\
\multicolumn{1}{l|}{conf\_thres=nfty} & \cellcolor[HTML]{FFFFFF}92.3\% & \cellcolor[HTML]{F7FCFA}92.6\% & \cellcolor[HTML]{F2FAF6}92.8\% & \cellcolor[HTML]{F2FAF6}92.8\% \\
 &
   &
   &
   &
   \\
\multicolumn{1}{c|}{\textbf{PEE}} &
  \cellcolor[HTML]{FFFFFF}num\_ensemble=2 &
  num\_ensemble=4 &
  num\_ensemble=6 &
  num\_ensemble=10 \\ \hline
\multicolumn{1}{l|}{conf\_thres=0.1} &
  \cellcolor[HTML]{FAE5E3}0.467 &
  \cellcolor[HTML]{F5C7C3}0.529 &
  \cellcolor[HTML]{F0AFAA}0.580 &
  \cellcolor[HTML]{E67C73}0.686 \\
\multicolumn{1}{l|}{conf\_thres=0.2} &
  \cellcolor[HTML]{FBE8E6}0.461 &
  \cellcolor[HTML]{FDF1F0}0.442 &
  \cellcolor[HTML]{FDF4F3}0.436 &
  \cellcolor[HTML]{FDF4F3}0.436 \\
\multicolumn{1}{l|}{conf\_thres=0.3} &
  \cellcolor[HTML]{FCEBE9}0.454 &
  \cellcolor[HTML]{FDF4F3}0.436 &
  \cellcolor[HTML]{FEFAF9}0.423 &
  \cellcolor[HTML]{FEFAF9}0.423 \\
\multicolumn{1}{l|}{conf\_thres=0.4} &
  \cellcolor[HTML]{FCEBE9}0.454 &
  \cellcolor[HTML]{FDF4F3}0.436 &
  \cellcolor[HTML]{FEFAF9}0.423 &
  \cellcolor[HTML]{FEFAF9}0.423 \\
\multicolumn{1}{l|}{conf\_thres=0.5} &
  \cellcolor[HTML]{FCEBE9}0.454 &
  \cellcolor[HTML]{FDF4F3}0.436 &
  \cellcolor[HTML]{FEFAF9}0.423 &
  \cellcolor[HTML]{FEFAF9}0.423 \\
\multicolumn{1}{l|}{conf\_thres=0.6} &
  \cellcolor[HTML]{FCEBE9}0.454 &
  \cellcolor[HTML]{FDF4F3}0.436 &
  \cellcolor[HTML]{FEFAF9}0.423 &
  \cellcolor[HTML]{FEF7F6}0.429 \\
\multicolumn{1}{l|}{conf\_thres=0.7} &
  \cellcolor[HTML]{FCEBE9}0.454 &
  \cellcolor[HTML]{FDF4F3}0.436 &
  \cellcolor[HTML]{FEFAF9}0.423 &
  \cellcolor[HTML]{FEFAF9}0.423 \\
\multicolumn{1}{l|}{conf\_thres=nfty} &
  \cellcolor[HTML]{FDF1F0}0.442 &
  \cellcolor[HTML]{FEFAF9}0.423 &
  \cellcolor[HTML]{FFFFFF}0.411 &
  \cellcolor[HTML]{FFFFFF}0.411 \\
 &
   &
   &
   &
   \\
\multicolumn{1}{c|}{\textbf{Valid Pct.}} &
  \cellcolor[HTML]{FFFFFF}num\_ensemble=2 &
  num\_ensemble=4 &
  num\_ensemble=6 &
  num\_ensemble=10 \\ \hline
\multicolumn{1}{l|}{conf\_thres=0.1} &
  \cellcolor[HTML]{B5E2CC}45.0\% &
  \cellcolor[HTML]{E5F5ED}17.4\% &
  \cellcolor[HTML]{FBFEFC}4.3\% &
  \cellcolor[HTML]{FFFFFF}1.6\% \\
\multicolumn{1}{l|}{conf\_thres=0.2} &
  \cellcolor[HTML]{90D2B2}66.8\% &
  \cellcolor[HTML]{A7DCC2}53.3\% &
  \cellcolor[HTML]{AEDEC6}49.6\% &
  \cellcolor[HTML]{B0E0C8}47.9\% \\
\multicolumn{1}{l|}{conf\_thres=0.3} &
  \cellcolor[HTML]{82CDA8}74.9\% &
  \cellcolor[HTML]{91D3B2}66.6\% &
  \cellcolor[HTML]{93D4B4}65.0\% &
  \cellcolor[HTML]{96D5B6}63.3\% \\
\multicolumn{1}{l|}{conf\_thres=0.4} &
  \cellcolor[HTML]{7CCAA4}78.8\% &
  \cellcolor[HTML]{87CFAC}72.2\% &
  \cellcolor[HTML]{8AD0AD}70.6\% &
  \cellcolor[HTML]{8DD1B0}68.8\% \\
\multicolumn{1}{l|}{conf\_thres=0.5} &
  \cellcolor[HTML]{78C9A1}81.1\% &
  \cellcolor[HTML]{82CDA8}75.1\% &
  \cellcolor[HTML]{85CEAA}73.4\% &
  \cellcolor[HTML]{88CFAC}71.5\% \\
\multicolumn{1}{l|}{conf\_thres=0.6} &
  \cellcolor[HTML]{75C79F}82.7\% &
  \cellcolor[HTML]{7FCBA6}77.0\% &
  \cellcolor[HTML]{82CDA8}75.3\% &
  \cellcolor[HTML]{85CEAA}73.6\% \\
\multicolumn{1}{l|}{conf\_thres=0.7} &
  \cellcolor[HTML]{73C79E}83.9\% &
  \cellcolor[HTML]{7CCAA4}78.5\% &
  \cellcolor[HTML]{7FCBA6}76.9\% &
  \cellcolor[HTML]{7FCCA6}76.8\% \\
\multicolumn{1}{l|}{conf\_thres=nfty} &
  \cellcolor[HTML]{57BB8A}100\% &
  \cellcolor[HTML]{57BB8A}100\% &
  \cellcolor[HTML]{57BB8A}100\% &
  \cellcolor[HTML]{57BB8A}100\%
\end{tabular}%
}
\caption{Performance of the estimator under varying ensemble sizes and uncertainty thresholds. Increasing the ensemble size generally enhances results, with 4 to 6 ensembles already providing sufficiently stable outcomes.}
\label{tab:ensemble_ablation}
\end{table}
\begin{table}[t]
\centering
\resizebox{\columnwidth}{!}{%
\begin{tabular}{l|cc}
 & \multicolumn{1}{l}{Inference Time (5 ensembles)} & \multicolumn{1}{l}{Max GPU Memory Allocation} \\ \hline
num\_events = 10k & 0.111 s & 0.70 GB \\
num\_events = 20k & 0.287 s & 1.36 GB \\
num\_events = 40k & 0.910 s & 2.71 GB \\
num\_events = 80k & 3.138 s & 5.39 GB
\end{tabular}%
}
\caption{Computational cost of our normal flow estimator.}
\label{tab:computation_cost}
\end{table}
\begin{table}[t]
\centering
\resizebox{\columnwidth}{!}{%
\begin{tabular}{@{}lccccc@{}}
\toprule
                            & \multicolumn{5}{c}{\#events every 20 ms -- quantile} \\ \cmidrule(l){2-6} 
                            & min     & 25\%       & 50\%     & 75\%      & max    \\ \midrule
MVSEC -- indoor\_flying1    & 85      & 2396.25    & 3720.0   & 5376.0    & 16177  \\
MVSEC -- indoor\_flying2    & 78      & 3157.0     & 5297.5   & 7987.5    & 23890  \\
MVSEC -- indoor\_flying3    & 78      & 2746.0     & 4850.0   & 6845.0    & 17476  \\
MVSEC -- outdoor\_day1      & 58      & 4412.0     & 6903.0   & 10646.75  & 96327  \\ \midrule
EVIMO -- IMO\_13\_00        & 7954    & 23564.5    & 35535.0  & 46824.5   & 68946  \\
EVIMO -- IMO\_13\_05        & 10806   & 55951.0    & 78963.0  & 87891.0   & 120730 \\
EVIMO -- SFM\_03\_00        & 2962    & 15766.0    & 76979.0  & 88460.5   & 105177 \\
EVIMO -- SFM\_03\_01        & 15315   & 41049.0    & 80223.0  & 94096.0   & 118441 \\ \midrule
DSEC -- interlaken\_00\_a   & 123449  & 133520.0   & 146397.0 & 157920.0  & 165356 \\
DSEC -- interlaken\_00\_b   & 183286  & 187122.75  & 189133.5 & 196445.75 & 209048 \\
DSEC -- thun\_01\_a         & 66476   & 83569.25   & 101858.5 & 117016.75 & 121001 \\
DSEC -- zurich\_city\_12\_a & 116746  & 138822.0   & 167154.0 & 196738.5  & 228827 \\ \bottomrule
\end{tabular}%
}
\vspace{-7pt}
\caption{Statistics of event density of each dataset.}
\label{tab:event_density}
\vspace{-12pt}
\end{table}
\begin{table*}[t]
\centering
\resizebox{\textwidth}{!}{%
\begin{tabular}{lccccccccccc}
\hline
\multicolumn{2}{l}{} &
   &
   &
  \multicolumn{2}{c}{Scene\_13\_00} &
  \multicolumn{2}{c}{Scene\_13\_05} &
  \multicolumn{2}{c}{Scene\_14\_03} &
  \multicolumn{2}{c}{Scene\_14\_04} \\ \cline{5-12} 
\multicolumn{2}{l}{\multirow{-2}{*}{}} &
  \multirow{-2}{*}{Input} &
  \multirow{-2}{*}{\begin{tabular}[c]{@{}c@{}}Training\\ Set\end{tabular}} &
  PEE $\downarrow$ &
  \% Pos $\uparrow$ &
  PEE $\downarrow$ &
  \% Pos $\uparrow$ &
  PEE $\downarrow$ &
  \% Pos $\uparrow$ &
  PEE $\downarrow$ &
  \% Pos $\uparrow$ \\ \hline
MultiCM &
  MB &
  F &
  - &
  \cellcolor[HTML]{E8847B}1.509 &
  \cellcolor[HTML]{FFFFFF}53.2\% &
  \cellcolor[HTML]{E67C73}4.315 &
  \cellcolor[HTML]{FFFFFF}75.7\% &
  \cellcolor[HTML]{E67C73}1.611 &
  \cellcolor[HTML]{FFFFFF}79.2\% &
  \cellcolor[HTML]{E67C73}1.800 &
  \cellcolor[HTML]{FFFFFF}73.2\% \\
PCA &
  MB &
  P &
  - &
  \cellcolor[HTML]{E67C73}1.573 &
  \cellcolor[HTML]{FFFFFF}88.2\% &
  \cellcolor[HTML]{F4C6C2}2.035 &
  \cellcolor[HTML]{FFFFFF}87.5\% &
  \cellcolor[HTML]{E78077}1.580 &
  \cellcolor[HTML]{ECF7F2}91.9\% &
  \cellcolor[HTML]{E77E75}1.784 &
  \cellcolor[HTML]{FCFEFD}90.3\% \\ \hline
 &
   &
   &
  M &
  \cellcolor[HTML]{EB948D}1.370 &
  \cellcolor[HTML]{FFFFFF}71.9\% &
  \cellcolor[HTML]{F2BAB5}2.406 &
  \cellcolor[HTML]{FAFDFB}90.6\% &
  \cellcolor[HTML]{EC9A93}1.356 &
  \cellcolor[HTML]{FFFFFF}69.5\% &
  \cellcolor[HTML]{ED9E97}1.458 &
  \cellcolor[HTML]{FFFFFF}64.6\% \\
\multirow{-2}{*}{E-RAFT} &
  \multirow{-2}{*}{SL} &
  \multirow{-2}{*}{F} &
  D &
  \cellcolor[HTML]{F7D3D0}0.843 &
  \cellcolor[HTML]{FFFFFF}{\ul 88.9\%} &
  \cellcolor[HTML]{FAE1DF}1.185 &
  \cellcolor[HTML]{B0DFC9}97.5\% &
  \cellcolor[HTML]{FFFBFB}{\ul 0.517} &
  \cellcolor[HTML]{FFFFFF}88.1\% &
  \cellcolor[HTML]{FEF8F8}{\ul 0.538} &
  \cellcolor[HTML]{FFFFFF}85.9\% \\ \hline
 &
   &
   &
  M &
  \cellcolor[HTML]{F7D5D2}0.823 &
  \cellcolor[HTML]{FFFFFF}85.6\% &
  \cellcolor[HTML]{EDA09A}3.201 &
  \cellcolor[HTML]{C8E9D9}95.3\% &
  \cellcolor[HTML]{F2B6B1}1.111 &
  \cellcolor[HTML]{FFFFFF}86.3\% &
  \cellcolor[HTML]{EC9790}1.532 &
  \cellcolor[HTML]{FFFFFF}86.0\% \\
\multirow{-2}{*}{TCM} &
  \multirow{-2}{*}{SSL} &
  \multirow{-2}{*}{F} &
  D &
  \cellcolor[HTML]{F9DBD8}{\ul 0.774} &
  \cellcolor[HTML]{FFFFFF}87.3\% &
  \cellcolor[HTML]{F1B5B0}2.541 &
  \cellcolor[HTML]{CAEADB}95.1\% &
  \cellcolor[HTML]{F7D2CF}0.872 &
  \cellcolor[HTML]{FFFFFF}87.8\% &
  \cellcolor[HTML]{F4C2BE}1.090 &
  \cellcolor[HTML]{FFFFFF}86.5\% \\ \hline
PointNet &
  SL &
  P &
  E &
  \cellcolor[HTML]{F2BBB6}1.047 &
  \cellcolor[HTML]{FFFFFF}88.1\% &
  \cellcolor[HTML]{FBE9E8}{\ul 0.924} &
  \cellcolor[HTML]{AFDFC8}{\ul 97.7\%} &
  \cellcolor[HTML]{F7D5D2}0.848 &
  \cellcolor[HTML]{9FD8BD}{\ul 98.3\%} &
  \cellcolor[HTML]{F8D6D3}0.892 &
  \cellcolor[HTML]{BEE5D3}{\ul 96.2\%} \\ \hline
 &
   &
   &
  M &
  \cellcolor[HTML]{FAE2E0}0.713 &
  \cellcolor[HTML]{C5E8D7}95.6\% &
  \cellcolor[HTML]{FFFEFE}0.269 &
  \cellcolor[HTML]{75C79F}99.3\% &
  \cellcolor[HTML]{FBE9E7}0.676 &
  \cellcolor[HTML]{8AD0AE}98.8\% &
  \cellcolor[HTML]{FCEDEC}0.651 &
  \cellcolor[HTML]{A7DCC3}98.1\% \\
 &
   &
   &
  D &
  \cellcolor[HTML]{FDF1F0}0.590 &
  \cellcolor[HTML]{BAE3D0}96.6\% &
  \cellcolor[HTML]{FFFFFF}\textbf{0.230} &
  \cellcolor[HTML]{60BF90}\textbf{99.8\%} &
  \cellcolor[HTML]{FDF4F4}0.575 &
  \cellcolor[HTML]{60BF90}\textbf{99.8\%} &
  \cellcolor[HTML]{FDF0EF}0.625 &
  \cellcolor[HTML]{6CC499}{\ul \textbf{99.5\%}} \\
 &
   &
   &
  E &
  \cellcolor[HTML]{FFFCFB}0.497 &
  \cellcolor[HTML]{B9E3CF}\textbf{96.7\%} &
  \cellcolor[HTML]{FEFAFA}0.399 &
  \cellcolor[HTML]{79C9A2}99.2\% &
  \cellcolor[HTML]{FFFFFF}\textbf{0.478} &
  \cellcolor[HTML]{79C9A2}99.2\% &
  \cellcolor[HTML]{FFFBFA}0.515 &
  \cellcolor[HTML]{8AD0AE}98.8\% \\
\multirow{-4}{*}{Ours} &
  \multirow{-4}{*}{SL} &
  \multirow{-4}{*}{P} &
  M+D+E &
  \cellcolor[HTML]{FFFFFF}\textbf{0.465} &
  \cellcolor[HTML]{BEE5D3}96.2\% &
  \cellcolor[HTML]{FFFDFD}0.308 &
  \cellcolor[HTML]{79C9A2}99.2\% &
  \cellcolor[HTML]{FEF8F7}0.544 &
  \cellcolor[HTML]{75C79F}99.3\% &
  \cellcolor[HTML]{FFFFFF}\textbf{0.467} &
  \cellcolor[HTML]{8AD0AE}98.8\% \\ \hline
 &
  \multicolumn{1}{l}{} &
  \multicolumn{1}{l}{} &
  \multicolumn{1}{l}{} &
   &
   &
   &
   &
   &
   &
   &
   \\ \hline
\multicolumn{3}{l}{} &
   &
  \multicolumn{2}{c}{scene\_14\_05} &
  \multicolumn{2}{c}{scene\_15\_01} &
  \multicolumn{2}{c}{scene\_15\_02} &
  \multicolumn{2}{c}{scene\_15\_05} \\ \cline{5-12} 
\multicolumn{3}{l}{\multirow{-2}{*}{}} &
  \multirow{-2}{*}{\begin{tabular}[c]{@{}c@{}}Training\\ Set\end{tabular}} &
  PEE $\downarrow$ &
  \% Pos $\uparrow$ &
  PEE $\downarrow$ &
  \% Pos $\uparrow$ &
  PEE $\downarrow$ &
  \% Pos $\uparrow$ &
  PEE $\downarrow$ &
  \% Pos $\uparrow$ \\ \hline
MultiCM &
  MB &
  F &
  - &
  \cellcolor[HTML]{E67C73}2.768 &
  \cellcolor[HTML]{FFFFFF}72.9\% &
  \cellcolor[HTML]{F5C7C4}0.852 &
  \cellcolor[HTML]{FFFFFF}68.0\% &
  \cellcolor[HTML]{F6CCC8}0.802 &
  \cellcolor[HTML]{FFFFFF}66.2\% &
  \cellcolor[HTML]{F7D4D1}0.744 &
  \cellcolor[HTML]{FFFFFF}59.8\% \\
PCA &
  MB &
  P &
  - &
  \cellcolor[HTML]{F1B5B0}1.823 &
  \cellcolor[HTML]{FFFFFF}89.4\% &
  \cellcolor[HTML]{E67C73}1.467 &
  \cellcolor[HTML]{E9F7F0}92.1\% &
  \cellcolor[HTML]{E67C73}1.612 &
  \cellcolor[HTML]{FFFFFF}78.2\% &
  \cellcolor[HTML]{E67C73}1.821 &
  \cellcolor[HTML]{FFFFFF}84.7\% \\ \hline
 &
   &
   &
  M &
  \cellcolor[HTML]{ED9F99}2.186 &
  \cellcolor[HTML]{FFFFFF}67.1\% &
  \cellcolor[HTML]{F4C2BD}0.899 &
  \cellcolor[HTML]{FFFFFF}72.7\% &
  \cellcolor[HTML]{F2BAB6}0.980 &
  \cellcolor[HTML]{FFFFFF}67.1\% &
  \cellcolor[HTML]{F2B7B2}1.100 &
  \cellcolor[HTML]{FFFFFF}57.9\% \\
\multirow{-2}{*}{ERAFT} &
  \multirow{-2}{*}{SL} &
  \multirow{-2}{*}{F} &
  D &
  \cellcolor[HTML]{FCEBEA}{\ul 0.908} &
  \cellcolor[HTML]{FFFFFF}86.3\% &
  \cellcolor[HTML]{FFFBFA}{\ul 0.432} &
  \cellcolor[HTML]{F1FAF6}91.3\% &
  \cellcolor[HTML]{FBE5E4}0.541 &
  \cellcolor[HTML]{F6FBF9}90.9\% &
  \cellcolor[HTML]{F8D9D7}{\ul 0.674} &
  \cellcolor[HTML]{FFFFFF}73.9\% \\ \hline
 &
   &
   &
  M &
  \cellcolor[HTML]{EA9088}2.445 &
  \cellcolor[HTML]{FFFFFF}82.2\% &
  \cellcolor[HTML]{FBE8E6}0.588 &
  \cellcolor[HTML]{FFFFFF}85.4\% &
  \cellcolor[HTML]{FAE4E2}0.556 &
  \cellcolor[HTML]{FFFFFF}87.7\% &
  \cellcolor[HTML]{F6CECB}0.811 &
  \cellcolor[HTML]{FFFFFF}68.0\% \\
\multirow{-2}{*}{TCM} &
  \multirow{-2}{*}{SSL} &
  \multirow{-2}{*}{F} &
  D &
  \cellcolor[HTML]{F3C0BB}1.640 &
  \cellcolor[HTML]{FFFFFF}84.1\% &
  \cellcolor[HTML]{FCEFEE}0.523 &
  \cellcolor[HTML]{FFFFFF}85.5\% &
  \cellcolor[HTML]{FBE7E5}{\ul 0.528} &
  \cellcolor[HTML]{FFFFFF}87.9\% &
  \cellcolor[HTML]{F5C9C6}0.871 &
  \cellcolor[HTML]{FFFFFF}68.1\% \\ \hline
PointNet &
  SL &
  P &
  E &
  \cellcolor[HTML]{FAE3E1}1.053 &
  \cellcolor[HTML]{BAE3D0}{\ul 96.6\%} &
  \cellcolor[HTML]{F7D2CF}0.765 &
  \cellcolor[HTML]{BFE6D3}{\ul 96.1\%} &
  \cellcolor[HTML]{F7D1CD}0.752 &
  \cellcolor[HTML]{CAEADB}{\ul 95.1\%} &
  \cellcolor[HTML]{F0B0AA}1.185 &
  \cellcolor[HTML]{F0F9F5}{\ul 91.5\%} \\ \hline
 &
   &
   &
  M &
  \cellcolor[HTML]{FDF1F0}0.806 &
  \cellcolor[HTML]{A3DAC0}98.2\% &
  \cellcolor[HTML]{FEF6F5}0.470 &
  \cellcolor[HTML]{BAE3D0}96.6\% &
  \cellcolor[HTML]{FDF0EF}0.433 &
  \cellcolor[HTML]{C3E7D5}95.8\% &
  \cellcolor[HTML]{FDF0EF}0.392 &
  \cellcolor[HTML]{D2EDE0}94.3\% \\
 &
   &
   &
  D &
  \cellcolor[HTML]{FFFFFF}{\ul \textbf{0.567}} &
  \cellcolor[HTML]{71C69C}{\ul \textbf{99.4\%}} &
  \cellcolor[HTML]{FFFFFF}0.391 &
  \cellcolor[HTML]{A7DCC3}{\ul \textbf{98.1\%}} &
  \cellcolor[HTML]{FFFDFD}0.298 &
  \cellcolor[HTML]{B5E1CC}{\ul \textbf{97.1\%}} &
  \cellcolor[HTML]{FCEDEC}0.424 &
  \cellcolor[HTML]{DEF2E8}93.2\% \\
 &
   &
   &
  E &
  \cellcolor[HTML]{FFFFFE}0.584 &
  \cellcolor[HTML]{92D3B4}98.6\% &
  {\ul \textbf{0.286}} &
  \cellcolor[HTML]{A7DCC3}{\ul \textbf{98.1\%}} &
  \cellcolor[HTML]{FFFFFF}{\ul \textbf{0.274}} &
  \cellcolor[HTML]{B8E3CE}96.8\% &
  \cellcolor[HTML]{FDF3F2}0.354 &
  \cellcolor[HTML]{C6E8D8}95.5\% \\
\multirow{-4}{*}{Ours} &
  \multirow{-4}{*}{SL} &
  \multirow{-4}{*}{P} &
  M+D+E &
  \cellcolor[HTML]{FFFFFF}0.568 &
  \cellcolor[HTML]{96D5B7}98.5\% &
  0.319 &
  \cellcolor[HTML]{AEDEC7}97.8\% &
  \cellcolor[HTML]{FFFDFD}0.300 &
  \cellcolor[HTML]{B6E2CD}97.0\% &
  \cellcolor[HTML]{FFFFFF}{\ul \textbf{0.201}} &
  \cellcolor[HTML]{C4E7D6}{\ul \textbf{95.7\%}} \\ \hline
\end{tabular}%
}
\caption{Per-scene normal flow evaluation on EVIMO2-\texttt{imo} split.}
\label{tab:evimo_per_scene_table}
\end{table*}

\end{document}